%% file: latex/ijcai26.tex
\documentclass{article}
\usepackage{ijcai26}

\usepackage{times}
\usepackage{soul}
\usepackage{url}
\usepackage[hidelinks]{hyperref}
\usepackage[utf8]{inputenc}
\usepackage[small]{caption}
\usepackage{graphicx}
\usepackage{amsmath}
\usepackage{amsthm}
\usepackage{booktabs}
\usepackage{algorithm}
\usepackage{algorithmic}
\usepackage[switch]{lineno}
\usepackage{natbib}
\usepackage{graphicx}
\usepackage{adjustbox}

\usepackage{hyperref}
\usepackage{url}
\usepackage{pifont}   
\usepackage{subcaption}
\usepackage{booktabs}       
\usepackage{tabularx}       
\usepackage{makecell}       
\usepackage{multirow}       
\usepackage{array}          
\usepackage{numprint}       
\usepackage{siunitx}
\usepackage{enumitem}       

\usepackage{booktabs} 
\usepackage{multirow} 
\usepackage{array} 
\usepackage{lscape}
\usepackage{longtable}
\usepackage{adjustbox} 
\usepackage{tabularx}

\usepackage{colortbl}
\usepackage{xcolor}
\usepackage[most]{tcolorbox}

\definecolor{rulegray}{gray}{0.85}

\usepackage{siunitx}
\title{Understanding Artificial Theory of Mind: Perturbed Tasks and Reasoning in Large Language Models}

\author{
    Christian Nickel$^{1,2,3}$
    \and
    Laura Schrewe$^{1}$\and
    Florian Mai$^{1,2}$\And
    Lucie Flek$^{1,2}$\\
    \affiliations
    $^1$Bonn-Aachen International Center for Information Technology (b-it), University of Bonn\\
    $^2$Lamarr Institute for Machine Learning and Artificial Intelligence\\
    $^3$Research Center Trustworthy Data Science and Security (RC-Trust), University of Duisburg-Essen
}

\begin{document}

\maketitle
\input{latex/content}

\clearpage

\section*{Ethical Considerations}
The dataset is entirely hand-written by the authors, without crowd-sourced labor. Sub-tasks are created by rule-based template instantiation (after \citeauthor{kosinskiTheoryMindMight2023}). While our work does not directly enable such applications, we note dual-use risk (manipulation/deception) and potential bias amplification in ToM-style reasoning; privacy concerns are mitigated because scenarios are fictional and contain no personal data.

\section*{Acknowledgements}
The authors gratefully acknowledge the granted access to the Bender cluster hosted by the University IT and Data Center (Hochschulrechenzentrum, HRZ) at the University of Bonn, which was essential for the computational experiments in this work.
This research was supported by the state of North Rhine-Westphalia as part of the Lamarr Institute for Machine Learning and Artificial Intelligence and the
Bonn-Aachen International Center for Information Technology (b-it), University of Bonn.

We thank Michal Kosinski (Stanford University) for making his dataset publicly available, which served as an important reference for our initial benchmark design.


\bibliographystyle{named}
\bibliography{latex/custom}

\appendix

\input{latex/appendix}

\end{document}

%% file: latex/content.tex
\begin{abstract}
Theory of Mind (ToM) refers to an agent's ability to model the internal states of others.
Contributing to the debate whether large language models (LLMs) exhibit genuine ToM capabilities, our study investigates their ToM robustness using perturbations on false-belief tasks and examines the potential of Chain-of-Thought prompting (CoT) to enhance performance and explain the LLM's decision. We introduce a handcrafted, richly annotated ToM dataset, including classic and perturbed false belief tasks, the corresponding spaces of valid reasoning chains for correct task completion, subsequent reasoning faithfulness, task solutions, and propose metrics to evaluate reasoning chain correctness and to what extent final answers are faithful to reasoning traces of the generated CoT. We show a steep drop in ToM capabilities under task perturbation for all evaluated LLMs, questioning the notion of any robust form of ToM being present. While CoT prompting improves the ToM performance overall in a faithful manner, it surprisingly degrades accuracy for some perturbation classes, indicating that selective application is necessary. 
\end{abstract}

\section{Introduction}

Theory of Mind (ToM) refers to an agent’s ability to infer and track the beliefs, intentions, and emotions of others \citep{premackDoesChimpanzeeHave1978, rabinowitz2018machine, kosinskiTheoryMindMight2023}. This ability to model the mental states of others is fundamental in human cognition and social interaction \citep{premackDoesChimpanzeeHave1978}. Hence, enabling reliable ToM abilities in AI agents could unlock a range of new applications involving human-AI interactions, e.g. in assistive healthcare~\citep{cuzzolin2020knowing, langley2022theory}, empathetic conversational agents~\citep{wang2024theory}, education~\citep{AsthanaCollinsThompson2024EducationalToM} or expert support, and cyber-physical systems like autonomous driving~\citep{montese2024policy}. 
Human ToM has been studied intensely in psychology and neuroscience, but the evidence of ToM in Large Language Models (LLMs) is mixed.
While promising results have been reported initially~\citep{kosinskiTheoryMindMight2023}, the underlying mechanisms remain unclear \citep{ullmanLargeLanguageModels2023}: Are LLMs truly reasoning about mental states, or merely leveraging statistical regularities?

Prior claims of ToM in LLMs often rely on narrow benchmarks that fail under small perturbations, calling into question their generality and interpretability. Existing benchmarks lack a systematic structure for evaluating these effects, and do not provide the means to isolate the impact of specific perturbation types or prompting strategies.

\begin{figure}[t]
\centering
\small
\setlength{\tabcolsep}{3pt}
\renewcommand{\arraystretch}{1.15}

\adjustbox{width=\columnwidth, frame, margin=4pt}{%
  \begin{minipage}{\linewidth} 
    
    \begin{tabularx}{\linewidth}{@{}lX@{}}
      \textbf{S1.} & \textcolor{blue!70!black}{The non-transparent bag contains sweets.} $\xrightarrow{\ \text{\footnotesize label}\ }$ \textcolor{green!70!black}{\texttt{unknown}}\\
      \textbf{S2.} & \textcolor{blue!70!black}{The bag’s label is \textquotedblleft vegetables\textquotedblright.} $\xrightarrow{\ \text{\footnotesize label}\ }$ \textcolor{green!70!black}{\texttt{vegetables}}\\
    \end{tabularx}
    
    \vspace{6pt} 

    \begin{tabularx}{\linewidth}{@{}l c c c@{}}
      \toprule
      \textbf{Model} & \textbf{After S1} & \textbf{After S2} & \textbf{Final output}\\
      \midrule
      No CoT           & \texttt{--}       & \texttt{--}        & \textcolor{red!80!black}{\texttt{sweets}}\\
      Incorrect CoT    & \textcolor{red!80!black}{\texttt{sweets}}   & \textcolor{green!70!black}{\texttt{vegetables}}& \textcolor{green!70!black}{\texttt{vegetables}}\\
      Correct CoT      & \textcolor{green!70!black}{\texttt{unknown}}  & \textcolor{green!70!black}{\texttt{vegetables}}& \textcolor{green!70!black}{\texttt{vegetables}}\\
      \bottomrule
    \end{tabularx}
    
  \end{minipage}%
}
\caption{Illustrative example of ToM task. We manually annotated every sentence with the correct current belief of a protagonist the agent has to reason about at each step. While CoT-P improves performance on some task classes, it degrades it on others. Our dataset allows to assess whether this is grounded in correct step-wise reasoning, where it fails and if models are faithful to their reasoning.}
\label{fig:minimal_example_compact}
\end{figure}

Our work addresses this gap by probing ToM robustness in LLMs through a novel dataset of systematically hand-crafted tasks and perturbations.
Each unperturbed task is accompanied by up to ten perturbed variants, constructed according to a diverse set of perturbation classes. This structure enables controlled comparisons and allows us to isolate the specific effects of different perturbation types, quantifying their impact on both performance and reasoning fidelity.

Additionally, we explore the effectiveness of Chain-of-Thought (CoT) prompting as a potential enhancement for ToM-related reasoning and robustness. To that end, we assess the CoTs’ faithfulness by examining whether their final outputs are truly grounded in intermediate reasoning steps. By annotating human-generated intermediate sets of valid gold CoT steps for every task instance, we introduce faithfulness metrics to evaluate whether final model predictions are grounded in intermediate gold-standard CoT steps, allowing for fine-grained evaluation of reasoning correctness. This analysis provides insight into how and when CoT prompting contributes to accurate reasoning or fails to do so.
Figure~\ref{fig:minimal_example_compact} illustrates this.

We use this benchmark to conduct a comparative evaluation of six open-source LLMs under both Vanilla Prompting (V-P) and Chain-of-Thought Prompting (CoT-P), assessing their performance on both standard and perturbed ToM tasks. Our results show that while task perturbation degrades the performance substantially, CoT prompting improves reasoning robustness in a faithful manner for most classes, but also reduces it for others.

In sum, these contributions provide a deeper understanding of the current limitations and opportunities for enhancing ToM-related capabilities in LLMs. 
The dataset, source code, and the model outputs of our experiments are available in the supplementary material and will be made available publicly to the research community upon acceptance.


\section{Related Work}

\paragraph{Theory of Mind in LLMs}
The emergence of ToM in LLMs has been explored in various works. \citet{kosinskiTheoryMindMight2023} argues that advanced models such as GPT-4 show signs of ToM, solving classical false-belief tasks at human-like accuracy levels. However, \citet{ullmanLargeLanguageModels2023} demonstrated that even minor perturbations disrupt performance, suggesting that models do not truly infer mental states but rather rely on statistical cues.
To further investigate the ToM capabilities of LLMs a variety of benchmarks have been created (see ~\citet{chen-etal-2025-theory} for a recent survey): ToMBench \citep{chenToMBenchBenchmarkingTheory2024a} provides a structured framework to evaluate ToM in LLMs, systematically covering all tasks and abilities that are considered part of ToM \citep{maHolisticLandscapeSituated2023}. They show that LLMs still struggle to comprehensively understand social scenarios and significantly lag behind human performance.
\citet{jonesEPITOMEExperimentalProtocol2023a} directly compare LLM and human performance on six experiments covering diverse ToM capabilities, showing that LLMs still make systematic errors in some tasks.
Many attempts have been made to broaden the scope and diversity of ToM evaluation benchmarks:
FANToM \citep{kimFANToMBenchmarkStresstesting2023} stresses models with dynamic multi-agent interactions, OpenToM \citep{xuOpenToMComprehensiveBenchmark2024} moves ToM assessment to complex social scenarios, ToMATO~\citep{shinoda2025tomato} is a large dataset generated from conversations between LLMs performing roleplay, and others evaluate ToM in a multi-modal setting~\citep{jin2024mmtom, shi2025muma}.

\citet{riemer-etal-2025-position} argue that most ToM benchmarks are fundamentally broken because they only measure \emph{literal} theory of mind, i.e. whether LLMs can predict the behavior of others, rather than whether the LLMs also adapt their own behavior accordingly (\emph{functional} theory of mind). For example, they find that LLMs struggle to adapt their \emph{Rock, Paper, Scissors} strategy to an opponent who only plays \emph{Rock}, despite being able to reliably predict their next move.
This behavior further supports the theory that LLMs rely to some extent on memorization or shortcuts to solve ToM tasks. 

To shed further light on this issue, our work systematically explores broad perturbation classes and their differential effects on ToM reasoning performance. Unlike existing benchmarks, our dataset provides interdependent gold CoT annotations across perturbed/unperturbed task pairs, allowing for a precise analysis of how perturbation types affect both reasoning quality and final predictions.

\paragraph{Chain-of-Thought}
\emph{Chain-of-Thought prompting} is a prompting technique designed to elicit intermediate reasoning steps from an LLM before it provides the final answer to a task~\citep{weiChainofThoughtPromptingElicits2022a}.
There are various strategies to elicit intermediate reasoning: Zero-shot prompting simply instructs the LLM to do so~\citep{kojimaLargeLanguageModels2023}, whereas the computationally more expensive few-shot prompting~\citep{brownLanguageModelsAre2020} provides few examples of the expected reasoning structure in the context to increase reliability. The present study uses one-shot prompting for a reasonable balance between speed and reliability\footnote{Some recent models were trained to produce intermediate reasoning via Reinforcement Learning~\citep{guo2025deepseek}. However, the structure of the resulting CoTs is generally not controllable, which makes these models unsuitable for our experimental design.}.

While Chain-of-Thought prompting improves performance on various reasoning benchmarks \citep{weiChainofThoughtPromptingElicits2022a}, its impact on Theory of Mind tasks remains uncertain and underexplored.
Some benchmarks report performance increase or decrease varying by task type \citep{xuOpenToMComprehensiveBenchmark2024}, while others report the performance impact to be small across all task types \citep{chenToMBenchBenchmarkingTheory2024a}.
Recently, a meta-analysis \citep{spragueCoTNotCoT2024} showed that CoT yields strong performance benefits primarily on tasks involving math or logic, with much smaller gains on other types of tasks, resulting from CoT improving symbolic execution.
While we also investigate the impact of CoT prompting, we focus on ToM tasks of the type of false belief. 

\paragraph{Chain-of-Though Faithfulness}
In addition to the potential performance impact, CoT prompting makes the LLMs produce a step-by-step explanation of their reasoning. This opens up possibilities for not treating LLMs as just a black box, thus benchmarking only question answering performance, but also to gain insights into the internal reasoning process and mistakes.
However, several results suggest that CoT reasoning traces are not faithful~\citep{turpinLanguageModelsDont2023} to the response and question the causal relevance of the produced CoT for the final answer~\citep{lanhamMeasuringFaithfulnessChainofThought2023}, demanding careful experimental designs to justify this attribution.

Similar in spirit to our work, \citet{jiang-etal-2025-matcha} employs a perturbation-based approach to test the robustness of general LLM reasoning.
Their findings that perturbations can induce inconsistent or nonsensical reasoning—especially on multi-step and commonsense tasks—complement our focus on measuring when CoT traces are informative versus misleading for ToM specifically.

Previous work on CoT faithfulness often relies on approximate measures such as ROUGE scores or structural similarity \citep{liFaithfulChainThoughtLarge2024}. 

To yield more precise faithfulness scores we use a correlation-based approach for measuring faithfulness that is based on the actual correctness of the reasoning chain and is thus more precise, while also computing several ROUGE-based approaches to model faithfulness, enabling us to compare the usefulness of the latter approximations to faithfulness.

\section{Dataset}
\begin{figure}[ht!]
    \centering
    \includegraphics[width=0.8\columnwidth]{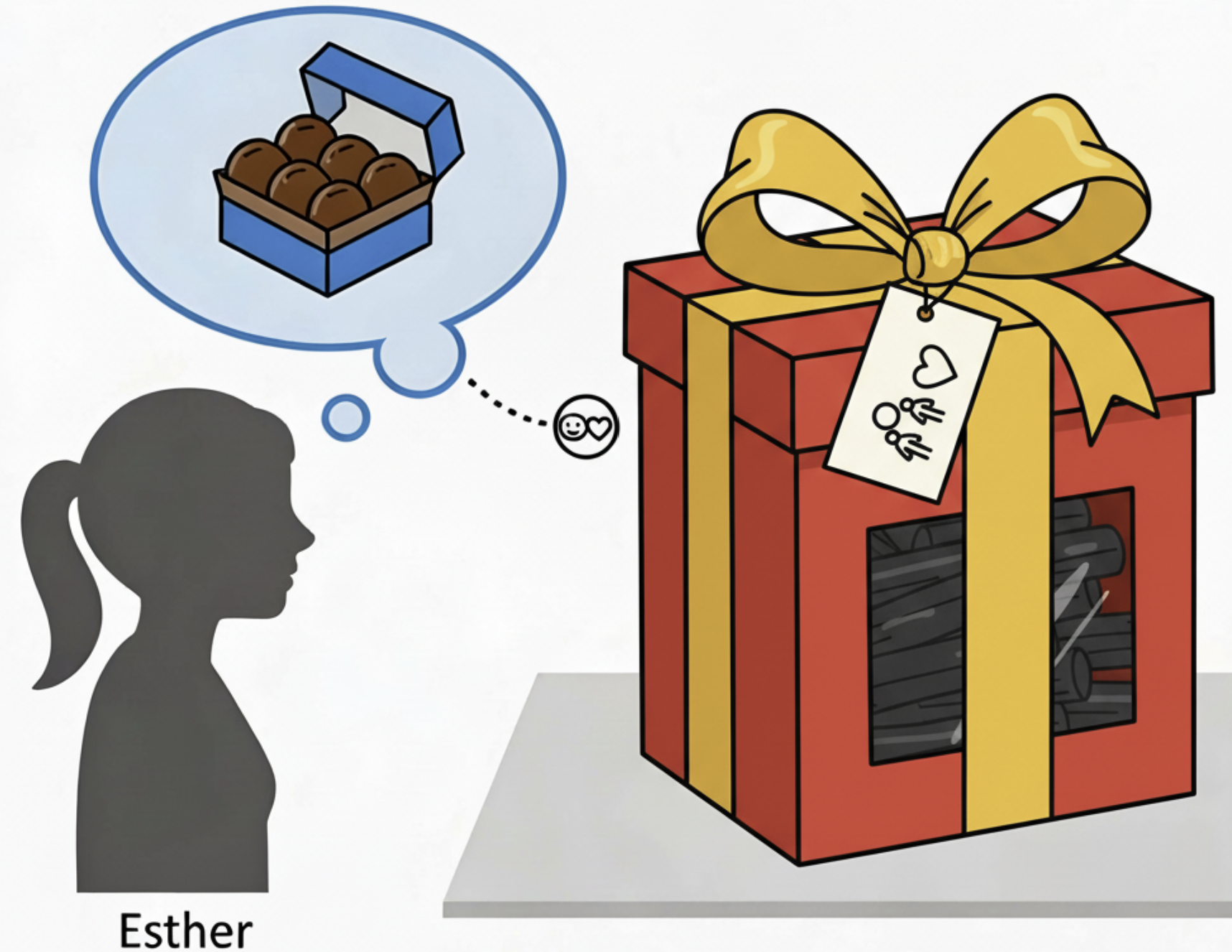} 
    \caption{Illustrating ``Conclusion from Sentiment''.}
    \label{fig:example-image-perturbed}
\end{figure}

\input{latex/dataset_florian}


\section{Experimental Setup}

\subsection{Evaluation Pipeline and Prompting Strategy}
We run inference with six open-source LLMs on our dataset under two prompting strategies: Vanilla Prompting (V-P), which directly requests an answer, and Chain-of-Thought Prompting (CoT-P), which elicits intermediate reasoning steps before answering.
Each model is instructed to return a structured JSON output containing mental state updates after each sentence in the task, followed by a final answer. This format enables automated evaluation of both answer accuracy and reasoning correctness (CoT). We use one-shot prompting with format demonstrations to improve output consistency, with minor adjustments across models. Faulty JSON outputs are filtered out.
After inference, we evaluate final answer accuracy, CoT correctness, and model faithfulness—defined as the statistical alignment between reasoning correctness and final predictions. To ensure reproducibility, details regarding our evaluation and analysis pipeline is provided in Supp.~\ref{sec:appendix:pipeline-figure}.

\subsection{Models}
We evaluate six recent open-source LLMs ranging from 33B to 132B parameters: Llama-2-70B-Chat \citep{touvronLlama2Open2023}, Llama-3-70B-Instruct \citep{IntroducingMetaLlamaa}, Vicuna-33B-v1.3 \citep{VicunaOpenSourceChatbot}, Yi-34B-Chat \citep{aiYiOpenFoundation2024}, Mixtral-8x7B-Instruct-v0.1 \citep{aiMixtralExperts2023}, and DBRX-Instruct \citep{IntroducingDBRXNew2024}.

All models are accessed via HuggingFace’s \texttt{transformers} library. Inference is run on A100-based compute nodes with the temperature set to 0. Further details, including hardware and inference parameters, are provided in Supp.~\ref{sec:appendix:implementation}.

\section{Evaluation Metrics and Definitions}

To evaluate the performance and reasoning capabilities of LLMs on Theory of Mind tasks, we rely on a multi-stage evaluation strategy.

\paragraph{Accuracy and Treatment Effects}
We first compute the accuracy of each model on false-belief tasks under both Vanilla Prompting (V-P) and Chain-of-Thought Prompting (CoT-P). To quantify the impact of perturbations and prompting strategies, we use \textit{Average Treatment Effect (ATE)} (i.e. the absolute difference), which allows us to measure how much model accuracy shifts when specific treatments (e.g., perturbations or prompting) are applied.

\paragraph{Robust Theory of Mind}
To determine whether a model demonstrates Theory of Mind (ToM) capabilities, we define a set of evaluation criteria based on accuracy thresholds. We consider a model to exhibit an \textbf{ostensible Theory of Mind (OToM)} if it achieves an accuracy above 50\% on unperturbed (i.e. classic) false-belief tasks. This baseline reflects performance exceeding random guessing in binary-choice scenarios.
To further assess the \textbf{robustness} of ToM, we define two additional criteria:
\begin{itemize}
    \item A model exhibits a \textbf{Robust Theory of Mind (RToM)} if it achieves $>$ 50\% accuracy \textit{on all ten} perturbation classes.
    \item A model exhibits a \textbf{Limited Robust Theory of Mind (limited RToM)} if it achieves $>$ 50\% accuracy on \textit{at least five} perturbation classes.
\end{itemize}
These thresholds allow us to distinguish between superficial ToM performance and more generalizable, perturbation-resilient reasoning abilities.

\paragraph{Identifying Easy and Hard Perturbation Classes}
We rank perturbation classes by their Average Treatment Effect (ATE) on model performance. To ensure robustness, we identify "challenging" classes via a set intersection of the top-five most degrading perturbations per model, using a majority incidence threshold ($4/6$ models) to still include perturbation classes where strict intersection is empty.

\paragraph{CoT Correctness}
For tasks evaluated under CoT-P, we assess the quality of the model-generated reasoning chains. Our primary metric is whether the predicted Chain of Thought (CoT) forms a \textit{proper subsequence} of one of the annotated gold-standard reasoning chains. 

According to our definition a \textbf{Proper Subsequence} must match valid intermediate belief states step-by-step and arrive at the correct final state, while allowing for some minor omissions such as skipped repetitions. In doing this we have to take into account path dependence as, given certain previous reasoning step states, only certain steps in the current step are allowed.  In the end this yields a binary output, determining whether we see a correct reasoning chain. It is the most precise measure as it tediously checks if the given CoT can be one of the valid chains
encoded in the gold rationale. We illustrate this in Figure~\ref{fig:propsubseq:examples}.
In addition to this binary metric, we also compute several continuous measures of CoT quality. They are approximative in nature, but easier to implement.
The first one is \textbf{Rouge-LCS (Longest Common Subsequence) based precision ($\text{ROUGE-LCS}_{P}$)} \citep{liFaithfulChainThoughtLarge2024}, which we adapt to our use case: 
       \begin{equation*}
        \text{ROUGE-LCS}_{P}=\frac{LCS(\text{Gold-CoT},\text{LLM-CoT})}{len(\text{LLM-CoT})}
        \label{formelzwei}
        \end{equation*} 

Employing (Pre-) Proper Subsequences, we define a Precision based on the \textbf{Longest Common (Pre-)Proper Subsequence}. We call it "pre-proper" as we have to drop the requirement that the last entries of both sequences have to match:
        \begin{equation*}
            \text{ROUGE-LCPS}_{P} =\frac{LCPS(\text{Gold-CoT},\text{LLM-CoT})}{len(\text{LLM-CoT})}
            \label{formeldrei}
        \end{equation*}
        where \text{LCPS}(X,Y) is the length of a longest common \textbf{pre-proper} subsequence of X and Y.

Last we define our metric \textbf{Transition Overlap Precision}:
    \begin{equation*}
     \text{Transition Overlap}_\text{Precision} = \frac{\|\text{GOLDSET} \cap \text{OUTPUTSET}\|}{\|    \text{OUTPUTSET}\|}
    \end{equation*}
where $\text{GOLDSET}$ is the set of all state transitions in the gold reasoning chain and $\text{OUTPUTSET}$ the set of all state transitions in the LLM generated reasoning.


\paragraph{Measuring Faithfulness}
We assess the \textbf{faithfulness} of CoT reasoning by computing the correlation between CoT correctness and final answer correctness. A high positive correlation suggests that the reasoning steps causally contribute to the final answer. We report $\Phi$-coefficients (for binary correctness based on proper subsequences) and point-biserial correlations ($r_{pb}$, for continuous CoT scores). A model is faithful in case of a positive, at least moderate to strong, positive correlation ($\phi \geq 0.4$, $r_{pb} \geq 0.4$ and $p \leq 0.05$).

\paragraph{Detecting Placebo Effects}
To distinguish meaningful CoT reasoning from superficial effects, we partition tasks and answers into two groups: those where the model generates a correct CoT, and those where it does not. If CoT-P substantially improves overall accuracy ($\text{ATE} \gg 0$) even given incorrect reasoning, we attribute this to a \textbf{placebo effect}, indicating that the structure or style of CoT prompts or other effects external to the actual ToM reasoning may alone influence outcomes.

\vspace{0.5em} 
Together, these analyses enable a fine-grained assessment of whether current LLMs exhibit reliable ToM capabilities, whether CoT prompting enhances those capabilities, and how these factors interact across models and task variations.

\input{latex/main-results-table}

\begin{table}[t]
\centering
\scriptsize
\setlength{\tabcolsep}{1.5pt}
\renewcommand{\arraystretch}{1.15}
\newcolumntype{Y}{>{\centering\arraybackslash}X}

\begin{tabularx}{\columnwidth}{l l Y Y Y Y Y Y}
\toprule
\makecell[l]{Statistic\\Type} & Measure &
\makecell[l]{Llama 2\\70B Chat} &
\makecell[l]{Vicuna\\33B v1.3} &
\makecell[l]{Mixtral\\8x7B Inst.} &
\makecell[l]{Yi\\34B Chat} &
\makecell[l]{Llama 3\\70B Inst.} &
\makecell[l]{DBRX\\Instruct} \\
\midrule
\multirow{5}{*}{\makecell[l]{Point-\\Biserial\\Correlation}}
& $\text{ROUGE-LCS}_{P}$                         & 0.391 & 0.516 & 0.304 & 0.245 & 0.398 & 0.235 \\
\arrayrulecolor{rulegray}\cline{2-8}\arrayrulecolor{black}
& \makecell[l]{$\text{ROUGE-LCPS}_{P}$} & 0.390 & 0.482 & 0.242 & 0.212 & 0.370 & 0.214 \\
\arrayrulecolor{rulegray}\cline{2-8}\arrayrulecolor{black}
& \makecell[l]{Transition\\Overlap}   & 0.476 & 0.501 & 0.281 & 0.229 & 0.435 & 0.224 \\
\midrule

\makecell[l]{Phi\\Coefficient}
& \makecell[l]{CoT\\Correctness} & 0.549 & 0.342 & 0.429 & 0.306 & 0.584 & 0.489 \\
\bottomrule
\end{tabularx}

\caption[Faithfulness: Correlations]{Faithfulness: Correlations between CoT-correctness and Final Answer Correctness for all models. Moderate to strong correlations suggest that models rely on their rationales, although other effects are present. P-values for these correlations are below the P = 0.05 threshold.}
\label{tab:RQ5:cot_correctness_correlations}
\end{table}

\section{Results and Analysis}
\subsection{Accuracies: ToM Robustness and Impact of Perturbations}
Without perturbations and using V-P, four of the evaluated models exhibited ostensible ToM-like behavior  (Table \ref{tab:accuracy_per_class_vanilla_vs_cot}). However, performance degrades under task perturbations, with only Llama-3-70B and DBRX maintaining limited robustness (limited RToM, $\geq$ 50\% accuracy across five perturbation classes). Especially perturbations that introduce the necessity of spatial reasoning pose challenges across all models.

The robustness to perturbations is higher with CoT-P (see Table~\ref{tab:accuracy_per_class_vanilla_vs_cot}), but Llama3-and DBRX remain the only models with limited robust ToM--now meeting the coin-toss threshold in 7 out of 10 perturbation classes. Nonetheless, spatial reasoning remains a consistent challenge across models, continuing to appear among the most impactful perturbation types.

\subsection{Effectiveness of CoT-P}
CoT-P slightly improved accuracy for unperturbed tasks and for all perturbation classes overall, but particularly \emph{Untrustworthy Testimony} and \emph{Add Unrelated Information}, as Table~\ref{tab:accuracy_per_class_vanilla_vs_cot} shows. Surprisingly, we even observe reduced performance in several perturbed tasks, especially \emph{Conclusion from Sentiment} and \emph{Automatic Change Knowledge}. Furthermore, we see decreases in accuracy for many sanity check tasks, which have a similar structure as false belief tasks, but do not test the Theory of Mind capabilities. This suggests that CoT-P introduces helpful structure, but if early details - like modifiers to the reasoning process or perceptive abilities of the protagonist (e.g. they cannot read) - are crucial for arriving at the correct reasoning steps and answer,  the task becomes challenging. 
This also happens when the model is asked for the true world state instead of the protagonist's belief state. This may be explained by the fact that the weight of intermediate information and reasoning steps with regards to the final answer decrease, leading to incorrect results. Thus, our results indicate that an application of CoT prompting techniques and improvements to the reasoning processes of the models ought to be selective. Moreover, perturbations affecting the spatial reasoning of a ToM task remain challenging regardless of the prompting technique.

\begin{figure*}[t]
	\includegraphics[width=\textwidth]{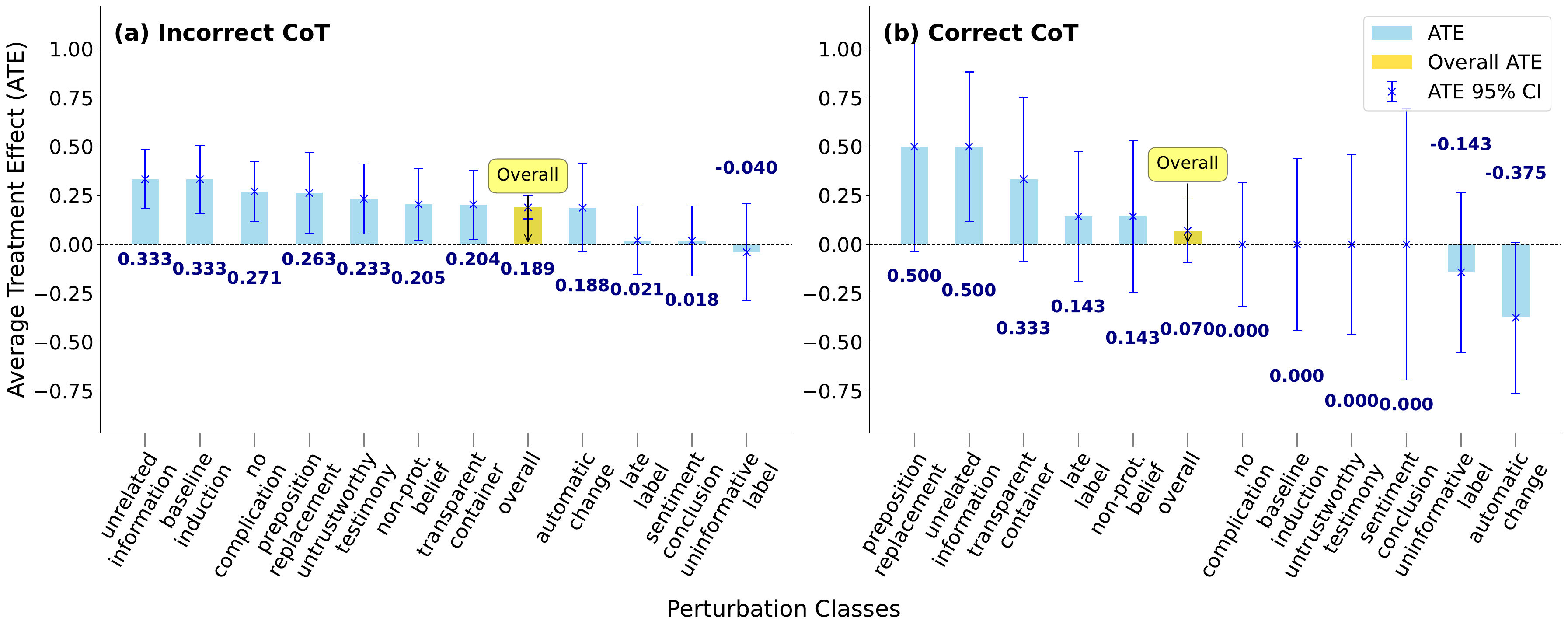}
	\caption[]{Comparison of effect strengths of CoT prompting given (a) incorrect (left-hand side) and (b) correct (right-hand side) reasoning rationales. Among all the evaluated models, we observe a placebo effect only in Mixtral, where incorrect reasoning (right) can even have a larger positive effect on final answer correctness than correct reasoning chains.}
 \label{fig:placebo:mixtral}
\end{figure*}

\subsection{Faithfulness of Final Answer to CoT}

In Table \ref{tab:RQ5:cot_correctness_correlations} we observe significant, mostly strong correlations between all measures of CoT- and final-answer - correctness, with it being strongest in larger and more recent models. We conclude models to be mostly faithful, though there also seem to be other influences present. When comparing the three approximative measures of faithfulness (Rouge-LCS, Rouge-LPS, and Transition Overlap) to the precise measure (Proper Subsequence), we note that they generally follow the same trends. Nevertheless, the correlations are usually smaller and there are a few outliers (inversion of the trend in Vicuna), indicating a limited utility of these simpler methods. 

\subsection{Placebo Effect}
The substantial positive effect of switching to CoT prompting can only be observed when the corresponding CoTs are correct - in most models. 
The only exception is Mixtral, where we witness a placebo effect (Figure \ref{fig:placebo:mixtral}). This is indicated by the improved performance under CoT-P, despite generating incorrect CoTs. A reason might be potential biases in reasoning steps.
Despite this negative example, in general we conclude that CoT correctness and reasoning play an important role in arriving at the correct final answer.



\section{Discussion}
\label{sec:discussion}

\paragraph{The Illusion of Robustness}
Our results lend empirical support to the skepticism regarding "emergent" Theory of Mind in LLMs. While models like Llama-3-70B and DBRX achieve high accuracy on classic false-belief tasks—often exceeding human baselines—this competence appears brittle. The steep performance drop observed under perturbations such as \textit{Transparent Container} and \textit{Preposition Replacement} suggests that these models rely on superficial heuristic matching (e.g., associating "looking inside" with "knowing") rather than maintaining a robust, generalized mental model of the agent. This aligns with \citet{ullmanLargeLanguageModels2023}'s hypothesis that current successes are likely instances of "ostensible" rather than "robust" Theory of Mind.

\paragraph{The Double-Edged Sword of CoT}
Contrary to the prevailing assumption that Chain-of-Thought (CoT) prompting universally enhances reasoning, our data reveals a more complex picture. 
While CoT proved beneficial for tasks from some perturbation classes, it unexpectedly degraded performance in other classes. 
This suggests that CoT can introduce noise or "reasoning hallucinations" when the underlying logic is not symbolic or math-heavy. Consequently, CoT should not be treated as a default solution for social reasoning tasks; its application must be selective and tailored to the specific complexity of the perturbation.

\paragraph{Faithfulness and Reasoning Fidelity}
A critical finding of our study is the strong positive correlation between the correctness of the generated reasoning traces (CoT) and the accuracy of the final answers, particularly in larger models. This implies a high degree of \textit{faithfulness}—models are generally not "making up" reasoning about ToM to justify a pre-determined guess, but rather fail because their intermediate reasoning steps are flawed. This distinction is vital: it suggests that improving the reasoning capabilities of LLMs (e.g., through better training data or guided reasoning) is a viable path toward robust ToM, as the models largely adhere to the logic they generate.

\input{latex/limitations_short}


\section{Conclusion}
We present a structured evaluation of Theory of Mind robustness in large language models, introducing a new ToM benchmark consisting of 10 perturbation classes and 1088 hand-crafted examples. Furthermore, we propose a faithfulness evaluation framework based on structured Chains of Thought, for which annotations were also created manually. Our findings show that four models demonstrate ostensible ToM on unperturbed tasks, but only Llama 3 70B and DBRX maintain limited robustness under perturbations. Spatial reasoning tasks persistently challenge all models, suggesting limitations in integrating ToM with spatial cognition. We find that Chain-of-Thought prompting can improve performance, particularly on unperturbed tasks, \emph{Untrustworthy Testimony}, and \emph{Add Unrelated Information}. However, CoT-P degrades performance in control tasks, \emph{Automatic Change Knowledge} and \emph{Conclusion from Sentiment}, indicating that it must be applied selectively. All evaluated models exhibit some degree of reasoning faithfulness, with CoT correctness positively correlated with final answer accuracy. Among tested CoT metrics, our novel metric based on proper subsequences offer the most precise evaluation of CoT correctness, while ROUGE-based approximations should be used with caution due to variability. Only one model showed placebo effects, where CoT structure alone improved performance; in general, correct CoTs are required for performance gains. Together, these findings highlight the fragility of ToM capabilities in LLMs, the promise and pitfalls of CoT prompting, and the need for rigorous evaluation tools. Our benchmark and framework offer a foundation for future research on robust and explainable ToM reasoning in LLMs. Ultimately, genuine ToM-like reasoning in LLMs remains fragile, but targeted prompting and structured evaluation offer a path forward. 


%% file: latex/dataset_florian.tex
\input{latex/example}

We introduce a novel, human-crafted dataset of tasks for evaluating ToM capabilities. The basic concept in these tasks follows the classic "False Belief" paradigm, testing an agent's ability to recognize and track beliefs of the protagonist about the world, which contradict reality.

The task dataset is organized into 7 \emph{stages}, which describes a scenario in which a task takes place. We newly created 4 "Unexpected Content" stages (popularized in the "Sally-Anne" test~\citep{wimmerBeliefsBeliefsRepresentation1983, baron1985does}) and 3 "Unexpected Transfer" (popularized in the "Smarties" test~\citep{perner1987three}) stages. In the former, a false belief is induced by changing the world state while the protagonist is absent. The latter, for which an example is given in Figure~\ref{fig:example}, induces false belief through a mislabeled container.
The task is to infer the \emph{protagonist's belief} about the state of the world.

\subsubsection{Perturbation Classes}

Given an unperturbed stage as the basis (see Figure~\ref{fig:example} for an example), we manually create a base task and up to 10\footnote{Not every stage is amenable to every perturbation.} alterations of this base task by introducing perturbations. The first 5 perturbation classes were already introduced by \citet{ullmanLargeLanguageModels2023}. However, they had not been systematically included and applied to a whole evaluation dataset. These classes are: 

\begin{enumerate}
  \item \textbf{Transparent Container:} The protagonist can see inside the container.
  \item \textbf{Preposition Replacement:} Changes spatial relations, e.g., "in" vs. "on".
  \item \textbf{Uninformative Label:} Protagonist cannot interpret the label.
  \item \textbf{Late Label:} The protagonist labeled or filled the container themselves.
  \item \textbf{Non-Protagonist Belief:} The question targets another agent's belief.
\end{enumerate}

Moreover, we introduce 5 novel perturbation classes:

\begin{enumerate}[resume]
  \item \textbf{Automatic Change Knowledge:} The object changes state automatically.
  \item \textbf{Add Unrelated Information:} Distractor details are introduced.
  \item \textbf{Induction from Baseline:} The protagonist infers based on past patterns.
  \item \textbf{Untrustworthy Testimony:} A known trickster gives misleading info.
  \item \textbf{Conclusion from Sentiment:} Beliefs are inferred from sentiment cues.
\end{enumerate}
We use these classes because they require different modes of reasoning to be integrated into the Theory of Mind process. First, some classes require spatial reasoning and understanding of transparency to understand what is perceivable by the protagonist (1 and 2). Then there are early modifiers that alter the protagonists historic knowledge about world states (4, 8, 10), behaviors and dynamics (6, 9) or ability to perceive or understand novel information (3) about the world. Lastly, there are classes that require simple filtering of noisy information (7) or taking the correct perspective (5).

An example of \textbf{Conclusion from Sentiment} is given in Figure~\ref{fig:example-perturbed} and illustrated in Figure \ref{fig:example-image-perturbed}. Examples for other classes are given in the supplementary material.

\subsubsection{Template-based Subtask Generation}

Finally, we automatically generate 16 alterations per task using the same template-based approach as \citet{kosinskiTheoryMindMight2023}, which comprise testing knowledge of the agent about the true world state, the belief of an informed protagonist, belief of a protagonist encountering an open container and the actual false belief of the protagonist. Moreover, the correct answers to the tasks are swapped. With 7 stages, up to 11 tasks per stage, and 16 subtasks per task, our dataset comprises a total of 1088 questions.

\subsubsection{Valid Reasoning Chains}

In order to answer a task question correctly, an agent needs to track the protagonist's belief as the scenario unfolds.
To analyze the correctness of CoT reasoning in the ToM false belief tasks, we manually annotated every task with the correct current state of the protagonist's belief at every step, where each step corresponds to one sentence in the task's scenario text.
Because sometimes the correct belief state is ambiguous based on the interpretation of the text, there can be multiple correct answers. Therefore, the correct current state is represented as a set. Example annotations are given in Figure~\ref{fig:dataset_inline_table}.

\begin{figure*}
    \centering
    \begin{subfigure}{0.49\textwidth}
        \centering
        \includegraphics[width=0.98\textwidth]{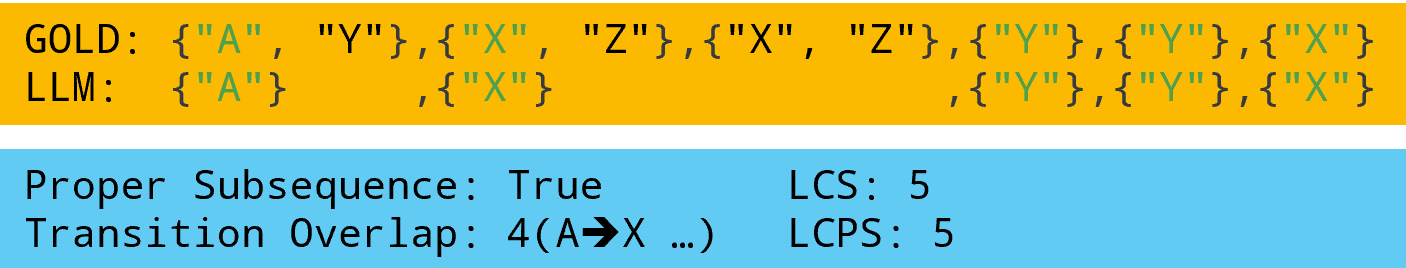}
        \caption{CoT of length $k=5$ is a proper subsequence of the solution.}
        \label{fig:propsubseq:valid}
    \end{subfigure}
    \hfill
    \begin{subfigure}{0.49\textwidth}
        \centering
        \includegraphics[width=0.98\textwidth]{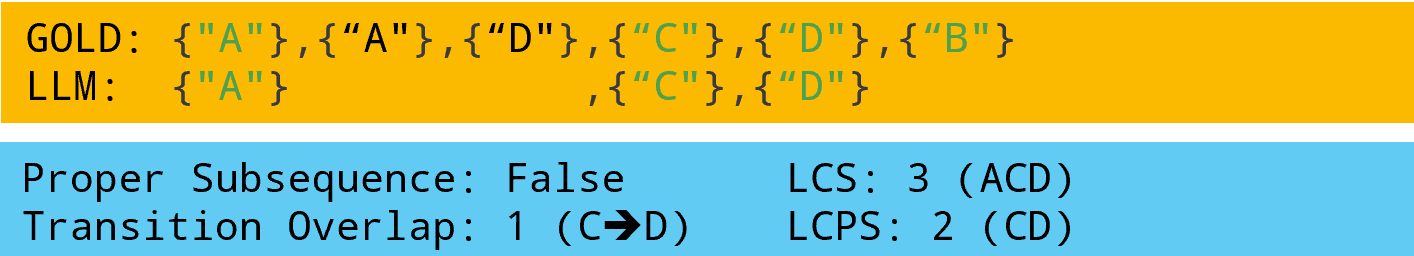}
        \caption{CoT of length $k=3$ is not a proper subsequence.}
        \label{fig:propsubseq:invalid}
    \end{subfigure}
    \caption[Examples of CoT reasoning correctness]{Examples of reasoning chains compared to the gold CoT. In (a), the model outputs a valid proper subsequence with consistent reasoning; in (b), intermediate states are inconsistent or skipped (step 3 "D" in the gold chain), leading to an invalid chain.}
    \label{fig:propsubseq:examples}
\end{figure*}

%% file: latex/example.tex
\begin{figure}[t!]
  \centering
  \captionsetup{font=small}

  \setlength{\tabcolsep}{4pt}
  \renewcommand{\arraystretch}{1.15}

  \begin{subfigure}[t]{\columnwidth}
    \centering
    \vspace{2pt}

    \adjustbox{width=\columnwidth,frame,margin=4pt}{
      \begin{tabularx}{\columnwidth}{@{}X l@{}}
        \toprule
        \textbf{Sentence} & \textbf{Belief} \\
        \midrule
        \textcolor{blue!70!black}{Esther finds a non-transparent paper bag she has never seen before.} & \textcolor{green!70!black}{\texttt{\{unknown\}}} \\
        \textcolor{blue!70!black}{In the paper bag are sweets and no vegetables.} & \textcolor{green!70!black}{\texttt{\{unknown\}}} \\
        \textcolor{blue!70!black}{The label on the paper bag says ``vegetables'' and not ``sweets''.} & \textcolor{green!70!black}{\texttt{\{unknown\}}} \\
        \textcolor{blue!70!black}{Esther does not open the paper bag.} & \textcolor{green!70!black}{\texttt{\{unknown\}}} \\
        \textcolor{blue!70!black}{She reads the label.} & \textcolor{green!70!black}{\texttt{\{vegetables\}}} \\
        \bottomrule
      \end{tabularx}
    }
        \caption{\textbf{Classic false-belief task (Unexpected Content)}}
        \label{fig:example}
  \end{subfigure}

  \vspace{6pt}

  \begin{subfigure}[t]{\columnwidth}
    \centering
    \vspace{2pt}

    \adjustbox{width=\columnwidth,frame,margin=4pt}{
      \begin{tabularx}{\columnwidth}{@{}X l@{}}
        \toprule
        \textbf{Sentence} & \textbf{Belief} \\
        \midrule
        \textcolor{blue!70!black}{Esther finds a non-transparent gift box in her room, with a gift card attached to it.} & \textcolor{green!70!black}{\texttt{\{unknown\}}} \\
        \textcolor{blue!70!black}{She does not know what is inside the gift box.} & \textcolor{green!70!black}{\texttt{\{unknown\}}} \\
        \textcolor{blue!70!black}{By shaking it she realizes that it contains small parts, like chocolate truffles or licorice.} & \textcolor{green!70!black}{\texttt{\{unknown\}}} \\
        \textcolor{blue!70!black}{She reads the gift card.} & \textcolor{green!70!black}{\texttt{\{unknown\}}} \\
        \textcolor{blue!70!black}{The text says the present is from her parents and that they are sure she will like the present.} & \textcolor{green!70!black}{\texttt{\{unknown\}}} \\
        \textcolor{blue!70!black}{Previously she mentioned to her parents that she really enjoys chocolate truffles and no other sweets or cookies, not even licorice.} & \textcolor{green!70!black}{\texttt{\{choc.\ truffles\}}} \\
        \textcolor{blue!70!black}{In the gift box are licorice and no chocolate truffles.} & \textcolor{green!70!black}{\texttt{\{choc.\ truffles\}}} \\
        \textcolor{blue!70!black}{Esther does not open the gift box.} & \textcolor{green!70!black}{\texttt{\{choc.\ truffles\}}} \\
        \bottomrule
      \end{tabularx}
    }
        \caption{\textbf{Perturbed task (Class: Conclusion from Sentiment)}}
        \label{fig:example-perturbed}
  \end{subfigure}

  \caption{Dataset illustration with per-sentence gold belief states shown inline.
  The belief column encodes what Esther believes about the container contents after each sentence.}
  \label{fig:dataset_inline_table}
\end{figure}

%% file: latex/main-results-table.tex
\begin{table*}[ht]
\scriptsize
\setlength{\tabcolsep}{2pt}
\renewcommand{\arraystretch}{1.2}
\newcolumntype{Y}{>{\centering\arraybackslash}X}

\newcommand{\U}[1]{\underline{#1}}              
\newcommand{\UB}[1]{\underline{\textbf{#1}}}    
\newcommand{\B}[1]{\textbf{#1}}                 

\newcommand{\pos}[1]{\textcolor{green!70!black}{#1}}
\newcommand{\negval}[1]{\textcolor{red!80!black}{#1}}

\begin{tabularx}{\textwidth}{l *{21}{Y}}
\toprule
\textbf{Perturbation Class}
& \multicolumn{3}{c}{\makecell{Llama 2\\70B Chat}}
& \multicolumn{3}{c}{\makecell{Vicuna\\33B v1.3}}
& \multicolumn{3}{c}{\makecell{Mixtral\\8$\times$7B Inst.}}
& \multicolumn{3}{c}{\makecell{Yi\\34B Chat}}
& \multicolumn{3}{c}{\makecell{Llama 3\\70B Inst.}}
& \multicolumn{3}{c}{\makecell{DBRX\\Instruct}}
& \multicolumn{3}{c}{\makecell{Average\\(all models)}} \\
\cmidrule(lr){2-4}\cmidrule(lr){5-7}\cmidrule(lr){8-10}\cmidrule(lr){11-13}\cmidrule(lr){14-16}\cmidrule(lr){17-19}\cmidrule(lr){20-22}
& \makecell[c]{V-P} & \makecell[c]{CoT} & \makecell[c]{ATE}
& \makecell[c]{V-P} & \makecell[c]{CoT} & \makecell[c]{ATE}
& \makecell[c]{V-P} & \makecell[c]{CoT} & \makecell[c]{ATE}
& \makecell[c]{V-P} & \makecell[c]{CoT} & \makecell[c]{ATE}
& \makecell[c]{V-P} & \makecell[c]{CoT} & \makecell[c]{ATE}
& \makecell[c]{V-P} & \makecell[c]{CoT} & \makecell[c]{ATE}
& \makecell[c]{V-P} & \makecell[c]{CoT} & \makecell[c]{ATE} \\
\midrule

No Perturbation
& \U{61.5} & \U{76.9} & \pos{+15.4}
& 27.3 & \U{81.8} & \pos{+54.6}
& \U{92.9} & \U{78.6} & \negval{-14.3}
& 42.9 & \UB{100.0} & \pos{+57.1}
& \U{100.0} & \U{100.0} & +0.0
& \U{78.6} & \U{100.0} & \pos{+21.4}
& \U{67.2} & \U{89.6} & \pos{+22.4} \\

Overall
& 46.6 & 46.6 & +0.0
& 42.7 & 49.6 & \pos{+6.8}
& 47.4 & \U{50.4} & \pos{+3.0}
& 46.5 & 48.8 & \pos{+2.3}
& \U{62.5} & \UB{69.9} & \pos{+7.4}
& \U{53.7} & \U{57.5} & \pos{+3.7}
& 49.9 & \U{53.8} & \pos{+3.9} \\

Transparent Container
& 28.6 & 28.6 & +0.0
& \UB{69.2} & \U{53.9} & \negval{-15.4}
& 7.1 & 35.7 & \pos{+28.6}
& 42.9 & 28.6 & \negval{-14.3}
& 42.9 & \U{57.1} & \pos{+14.3}
& 42.9 & \U{57.1} & \pos{+14.3}
& 38.9 & 43.5 & \pos{+4.6} \\

Preposition Replacement
& 20.0 & 0.0 & \negval{-20.0}
& 40.0 & 30.0 & \negval{-10.0}
& 20.0 & 20.0 & +0.0
& \UB{60.0} & 30.0 & \negval{-30.0}
& 20.0 & 40.0 & \pos{+20.0}
& 37.5 & 25.0 & \negval{-12.5}
& 32.9 & 24.2 & \negval{-8.8} \\

Uninformative Label
& 12.5 & 0.0 & \negval{-12.5}
& 37.5 & 25.0 & \negval{-12.5}
& 50.0 & 25.0 & \negval{-25.0}
& 25.0 & 12.5 & \negval{-12.5}
& 25.0 & 37.5 & \pos{+12.5}
& \UB{62.5} & \U{62.5} & +0.0
& 35.4 & 27.1 & \negval{-8.3} \\

Late Label
& \U{53.9} & \U{61.5} & \pos{+7.7}
& 33.3 & \UB{75.0} & \pos{+41.7}
& 42.9 & 42.9 & +0.0
& 38.5 & 23.1 & \negval{-15.4}
& 42.9 & 42.9 & +0.0
& \U{64.3} & \U{57.1} & \negval{-7.2}
& 45.9 & \U{50.4} & \pos{+4.5} \\

Non Protagonist Belief
& \U{64.3} & \U{71.4} & \pos{+7.1}
& \U{66.7} & \U{77.8} & \pos{+11.1}
& \U{69.2} & \U{84.6} & \pos{+15.4}
& \U{57.1} & 50.0 & \negval{-7.1}
& \UB{92.9} & \U{92.9} & +0.0
& \U{78.6} & \U{85.7} & \pos{+7.1}
& \U{71.5} & \U{77.1} & \pos{+5.6} \\

Automatic Change Knowledge
& \U{60.0} & 40.0 & \negval{-20.0}
& 37.5 & 25.0 & \negval{-12.5}
& \U{60.0} & 30.0 & \negval{-30.0}
& \UB{66.7} & \U{66.7} & +0.0
& 50.0 & 50.0 & +0.0
& \U{60.0} & 50.0 & \negval{-10.0}
& \U{55.7} & 43.6 & \negval{-12.1} \\

Add Unrelated Information
& \U{57.1} & \U{85.7} & \pos{+28.6}
& 50.0 & 35.7 & \negval{-14.3}
& \U{64.3} & \UB{100.0} & \pos{+35.7}
& \U{64.3} & \U{64.3} & +0.0
& \U{100.0} & \U{100.0} & +0.0
& \U{64.3} & \U{85.7} & \pos{+21.4}
& \U{66.7} & \U{78.6} & \pos{+11.9} \\

Induction From Baseline
& \U{54.6} & 36.4 & \negval{-18.2}
& \U{63.6} & 45.5 & \negval{-18.2}
& 50.0 & 50.0 & +0.0
& 36.4 & \U{54.6} & \pos{+18.2}
& \U{66.7} & \UB{91.7} & \pos{+25.0}
& 33.3 & 33.3 & +0.0
& \U{50.8} & \U{51.9} & \pos{+1.1} \\

Untrustworthy Testimony
& 33.3 & \U{58.3} & \pos{+25.0}
& 0.0 & 50.0 & \pos{+50.0}
& 50.0 & \U{66.7} & \pos{+16.7}
& 33.3 & \U{66.7} & \pos{+33.3}
& \U{66.7} & \UB{75.0} & \pos{+8.3}
& 33.3 & 50.0 & \pos{+16.7}
& 36.1 & \U{61.1} & \pos{+25.0} \\

Conclusion From Sentiment
& 50.0 & 21.4 & \negval{-28.6}
& 36.4 & 36.4 & +0.0
& 14.3 & 0.0 & \negval{-14.3}
& 46.2 & 30.8 & \negval{-15.4}
& 50.0 & \UB{57.1} & \pos{+7.1}
& 28.6 & 7.1 & \negval{-21.4}
& 37.6 & 25.5 & \negval{-12.1} \\

\bottomrule
\end{tabularx}
\caption[Accuracy by Perturbation Class, Vanilla prompting (V-P) vs Chain-of-Thought (CoT)]{Accuracy (as percentages \%) per perturbation class and model, comparing Vanilla prompting (V-P) vs Chain-of-Thought (CoT). \U{Underline} denotes performance above 50\%, and \textbf{bold} denotes the best model performance on the data subset. ATE reports the signed absolute change (CoT$-$V-P), with \textcolor{green!70!black}{improvements in green} and \textcolor{red!80!black}{deteriorations in red}. We additionally report the \textbf{Average} over all models.}
\label{tab:accuracy_per_class_vanilla_vs_cot}
\end{table*}

%% file: latex/limitations_short.tex
\subsubsection{Limitations and Future Work}
While our work provides a structured investigation into the Theory of Mind robustness and reasoning faithfulness in LLMs, several limitations remain. First, our primary metrics for CoT correctness rely on structured reasoning outputs, which smaller or less instruction-tuned models often fail to produce consistently. This restricts our ability to evaluate such models using our most precise metrics. However, due to the rapid improvement in instruction following even in very small models, we do not expect this to be an issue in the future.
Second, the dataset is focused solely on false belief tasks, leaving out other important Theory of Mind dimensions such as desire reasoning or faux pas detection. Broadening the task diversity could yield deeper insights into the social cognition of LLMs. However, applying a manual data creation process as in the present is likely too expensive to scale to large datasets. Hence, future work may resort to LLMs for synthetic data generation.
Third, our current evaluation and analysis is limited to open-source models, excluding leading proprietary models such as GPT, Claude, or Gemini to ensure reproducibility. Furthermore, gathering new human performance data on the perturbed tasks would provide critical baselines and allow for more meaningful comparisons between LLMs and human reasoning and how far current models deviate from human-level performance and understanding in these scenarios.
Finally, exploring alternative prompting approaches such as SIMTOM \citep{wilfThinkTwicePerspectiveTaking2023}, which filters context into protagonist-perspective views, agent-based architectures, and symbolic approaches that explicitly model narrator and character states separately, could lead to more robust and interpretable ToM in LLMs, which could be in turn investigated using our dataset and approach in this work. 

%% file: latex/appendix.tex
\clearpage

\section*{Supplementary Material}

\section{Approach Details}
\label{sec:appendix:approach}

Our experimental setup is structured around four core components: a novel dataset, multiple prompting configurations, targeted evaluation metrics, and a multi-model inference pipeline. Each step is designed to support analysis of Theory of Mind (ToM) robustness and reasoning quality in LLMs. We analyze model behavior under both unperturbed and systematically perturbed false-belief tasks.

The reasoning chains (Chains of Thought, CoTs) are evaluated against manually annotated gold rationales. These allow us to measure not only final-answer correctness but also the quality and faithfulness of model-generated reasoning steps.


\section{Perturbation Classes and Examples}
\label{sec:appendix:complexity-classes}

We define 10 perturbation classes to test ToM robustness. Five are adapted from \citet{ullmanLargeLanguageModels2023}; five are newly introduced in this work.

\subsection*{Original Perturbation Classes (from Ullman)}
\begin{itemize}
  \item \textbf{Transparent Container:} The protagonist can see inside the container.
  \begin{quote}\small
  "Esther finds a transparent bag. She can see that it contains sweets."
  \end{quote}

  \item \textbf{Preposition Replacement:} Changes spatial relations, e.g., "in" vs. "on".
  \begin{quote}\small
  "Charlie finds a non-transparent box. On the box are mangos."
  \end{quote}

  \item \textbf{Uninformative Label:} Protagonist cannot interpret the label.
  \begin{quote}\small
  "Charlie finds a box labeled 'bananas' but cannot read."
  \end{quote}

  \item \textbf{Late Label:} The protagonist labeled or filled the container themselves.
  \begin{quote}\small
  "Charlie filled the box with mangos yesterday. The label says 'bananas'."
  \end{quote}

  \item \textbf{Non-Protagonist Belief:} The question targets another agent's belief.
  \begin{quote}\small
  "What does Carl believe about the location of the ice cream?"
  \end{quote}
\end{itemize}

\subsection*{New Perturbation Classes (introduced in this work)}
\begin{itemize}
  \item \textbf{Automatic Change Knowledge:} The object changes state automatically.
  \begin{quote}\small
  "Charlie buys green mangos. Mangos ripen and turn red. Charlie does not know this."
  \end{quote}

  \item \textbf{Add Unrelated Information:} Distractor details are introduced.
  \begin{quote}\small
  "Aya finds a bottle. Next to it are truffles, candy boxes, and museum artifacts."
  \end{quote}

  \item \textbf{Induction from Baseline:} The protagonist infers based on past patterns.
  \begin{quote}\small
  "Esther assumes the bag contains sweets because it always does."
  \end{quote}

  \item \textbf{Untrustworthy Testimony:} A known trickster gives misleading info.
  \begin{quote}\small
  "Her sister says it's sweet sauce, but usually lies."
  \end{quote}

  \item \textbf{Conclusion from Sentiment:} Beliefs are inferred from sentiment cues.
  \begin{quote}\small
  "The gift is from her parents, who know she only likes licorice."
  \end{quote}
\end{itemize}


\section{Definitions and Evaluation Metrics}
\label{sec:appendix:definitions}

\subsection*{Accuracy}
Given a binary classification setup (correct vs. incorrect final answer):
\begin{equation}
\text{Accuracy} = \frac{\text{Correct Final Answers}}{\text{Total Final Answers}}
\end{equation}

\subsection*{Average Treatment Effect (ATE)}
To measure the effect of a treatment (e.g., CoT prompting or perturbation):
\begin{equation}
\text{ATE} = P(Y=1 | T=1) - P(Y=1 | T=0)
\end{equation}

\subsection*{Relative Risk (RR)}
To express how many times more likely a correct answer is under treatment:
\begin{equation}
\text{RR} = \frac{P(Y=1 | T=1)}{P(Y=1 | T=0)}
\end{equation}

\subsection*{CoT- Correctness}
\paragraph{Proper Subsequence}
We introduce a \textbf{Proper Subsequence} is a sequence where each element of the generated sequence matches the corresponding element in the reference sequence, while still allowing for the skipping of repeated entries in the reference sequence. Additionally, the final elements of both sequences must match.

The algorithm can be found in Listing \ref{lst:propersubsequence}. 

\lstset{
    backgroundcolor=\color{lightgray},
    frame=single,
    basicstyle=\footnotesize\ttfamily,
    numbers=left,
    numberstyle=\tiny,
    numbersep=5pt,
    keywordstyle=\color{blue},
    commentstyle=\color{yellow},
    stringstyle=\color{red},
    breaklines=true,
    breakatwhitespace=true
}
\begin{lstlisting}[float=t, frame=single,language=Python,     captionpos=b, caption={Algorithm to determine if a candidate sequence is a proper subsequence of a gold sequence.}, label=lst:propersubsequence]
def is_proper_subsequence(LLM-CoT, Gold-CoT):
i = 0
j = 0
while i < len(LLM-CoT) and j < len(Gold-CoT):
	if LLM-CoT[i] == Gold-CoT[j]: 
		i += 1
		j += 1
	elif Gold-CoT[j-1] == Gold-CoT[j] and j<len(Gold-CoT)-1: 
		#skip duplicate / look ahead
		j += 1
	else:
		return False
if i == len(LLM-CoT) and LLM-CoT[-1]==Gold-CoT[-1]:
	return True
\end{lstlisting}

\subsection*{Faithfulness}
Faithfulness is computed as the correlation between CoT correctness and final answer correctness. For binary CoT correctness (e.g., proper subsequence match), we use the Phi-coefficient:

The Phi Coefficient (\(\phi\)) is a measure of association for two dichotomous variables. It is calculated as:
\[
\phi = \frac{n_{11}n_{00} - n_{10}n_{01}}{\sqrt{(n_{1\cdot}n_{0\cdot}n_{\cdot1}n_{\cdot0})}}
\]
where:
\begin{itemize}
    \item \(n_{11}\) is the number of cases where both variables are 1,
    \item \(n_{00}\) is the number of cases where both variables are 0,
    \item \(n_{10}\) is the number of cases where the first variable is 1 and the second is 0,
    \item \(n_{01}\) is the number of cases where the first variable is 0 and the second is 1,
    \item \(n_{1\cdot}\) is the total number of cases where the first variable is 1,
    \item \(n_{0\cdot}\) is the total number of cases where the first variable is 0,
    \item \(n_{\cdot1}\) is the total number of cases where the second variable is 1,
    \item \(n_{\cdot0}\) is the total number of cases where the second variable is 0.
\end{itemize}
\hfill  \citep{sheskinHandbookParametricNonparametric2020}

For continuous CoT scores (e.g., ROUGE-L), we use the point-biserial correlation:
    \begin{equation}
    r_{pb} = \frac{M_1 - M_0}{s_n} \sqrt{\frac{n_1 n_0}{n(n-1)}}
    \end{equation}
    Where:
    \begin{itemize}
    \item $r_{pb}$ is the point-biserial correlation coefficient
    \item $M_1$ is the mean of the continuous variable for the group coded as 1
    \item $M_0$ is the mean of the continuous variable for the group coded as 0
    \item $s_n$ is the standard deviation of the continuous variable for the entire sample
    \item $n_1$ is the number of cases in the group coded as 1
    \item $n_0$ is the number of cases in the group coded as 0
    \item $n$ is the total sample size
    \end{itemize}
    \hfill \citep{glassStatisticalMethodsEducation1996}


\section{Implementation Details}
\label{sec:appendix:implementation}

We run inference using Python 3.9 and Huggingface Transformers. Model responses are collected via batch inference on A100 GPUs.
We test the following models:
\begin{itemize}
  \item LLaMA 2 70B Chat
  \item Vicuna 33B v1.3
  \item Yi-34B-Chat
  \item Mixtral 8x7B Instruct
  \item LLaMA 3 70B Instruct
  \item DBRX Instruct
\end{itemize}

All models are evaluated using both Vanilla Prompting (V-P) and Chain-of-Thought Prompting (CoT-P). In CoT-P, output is expected in JSON format with intermediate mind states and a final answer. Prompts and parsing logic are designed to support structured evaluation and CoT correctness scoring. Full prompt templates are included in the supplementary material.


\section{Data and Code Availability}
\label{sec:appendix:availability}

To prevent data contamination, we will release the dataset and annotation files for academic purposes through a gated access via e.g. HuggingFace Datasets. If interested in the dataset right now, researchers may request access for non-commercial academic use by contacting the corresponding author. 
We support reproducibility and open science, and can provide the scripts for template based generation, inference, analysis, the dataset, gold labels, prompts, and scoring functions model outputs, CoT chains, evaluation metrics, and analysis notebooks upon request.

\onecolumn

\section{Supplementary Tables}
\label{sec:appendix:tables}

Here we provide full breakdowns of our evaluation results.

\subsection{Accuracies}
(see next page for full table)
\input{tables/RQ3/accuracies}

\subsection{ATE with Perturbations as Treatment, All-Subtasks}
(see next page for full table)
\input{tables/RQ3/effect_strengths_pert_treatment_all_subtasks}

\subsection{ATE with Perturbations as Treatment, False Belief Tasks}
(see next page for full table)
\input{tables/RQ3/effect_strengths_pert_treatment_false_belief}

\subsection{ATE with CoT-P as Treatment}
(see next page for full table)
\input{tables/RQ3/effect_strengths_cottreatment}

\subsection{Faithfulness}
\begin{table*}

\setlength{\tabcolsep}{1.5pt}
\newcolumntype{Y}{>{\centering\arraybackslash}X}

\begin{scriptsize}
\resizebox{\textwidth}{!}{%
\begin{tabularx}{\textwidth}{l l l Y Y Y Y Y Y}
\toprule
\makecell[l]{Statistic\\Type} & Measure & Metric &
\makecell[l]{Llama 2\\70B Chat} &
\makecell[l]{Vicuna\\33B v1.3} &
\makecell[l]{Mixtral\\8x7B Inst.} &
\makecell[l]{Yi\\34B Chat} &
\makecell[l]{Llama 3\\70B Inst.} &
\makecell[l]{DBRX\\Instruct} \\
\midrule

\multirow{6}{*}{\makecell[l]{Point-\\Biserial\\Correlation}} 
& \multirow{2}{*}{ROUGE-L} & Corr. & 0.391 & 0.516 & 0.304 & 0.245 & 0.398 & 0.235 \\
& & P-Value & \num{2e-6} & 0 & \num{3.14e-4} & \num{4e-3} & \num{2e-6} & \num{5.989e-3} \\
\cmidrule{2-9}

& \multirow{2}{*}{\makecell[l]{ROUGE-L\\(Pre-Proper)}} & Corr. & 0.39 & 0.482 & 0.242 & 0.212 & 0.37 & 0.214 \\
& & P-Value & \num{3e-6} & 0 & \num{4.613e-3} & \num{1.3389e-2} & \num{1e-5} & \num{1.2283e-2} \\
\cmidrule{2-9}

& \multirow{2}{*}{\makecell[l]{Transition\\Overlap}} & Corr. & 0.476 & 0.501 & 0.281 & 0.229 & 0.435 & 0.224 \\
& & P-Value & 0 & 0 & \num{9.35e-4} & \num{7.429e-3} & 0 & \num{8.75e-3} \\

\midrule

\multirow{2}{*}{\makecell[l]{Phi\\Coefficient}} 
& \multirow{2}{*}{\makecell[l]{CoT\\Correctness}} & Corr. & 0.549 & 0.342 & 0.429 & 0.306 & 0.584 & 0.489 \\
& & P-Value & 0 & \num{6.7e-5} & \num{1e-6} & \num{3.54e-4} & 0 & 0 \\

\bottomrule
\end{tabularx}
}
\end{scriptsize}
\caption[Faithfulness: Correlations and P-Values]{Faithfulness: Correlations and P-Values between CoT-correctness and Final Answer Correctness for all models. Moderate to strong correlations suggest that models rely on their rationales, although other effects are present.}
\label{tab:RQ5:cot_correctness_correlations}

\end{table*}


\section{Supplementary Figures}
For a more intuitive understanding of both the dataset and the evaluation pipeline we provide some visual representations.

\subsection{Dataset Structure}
\begin{figure*}[h]
	\centering
	\includegraphics[width=\textwidth]{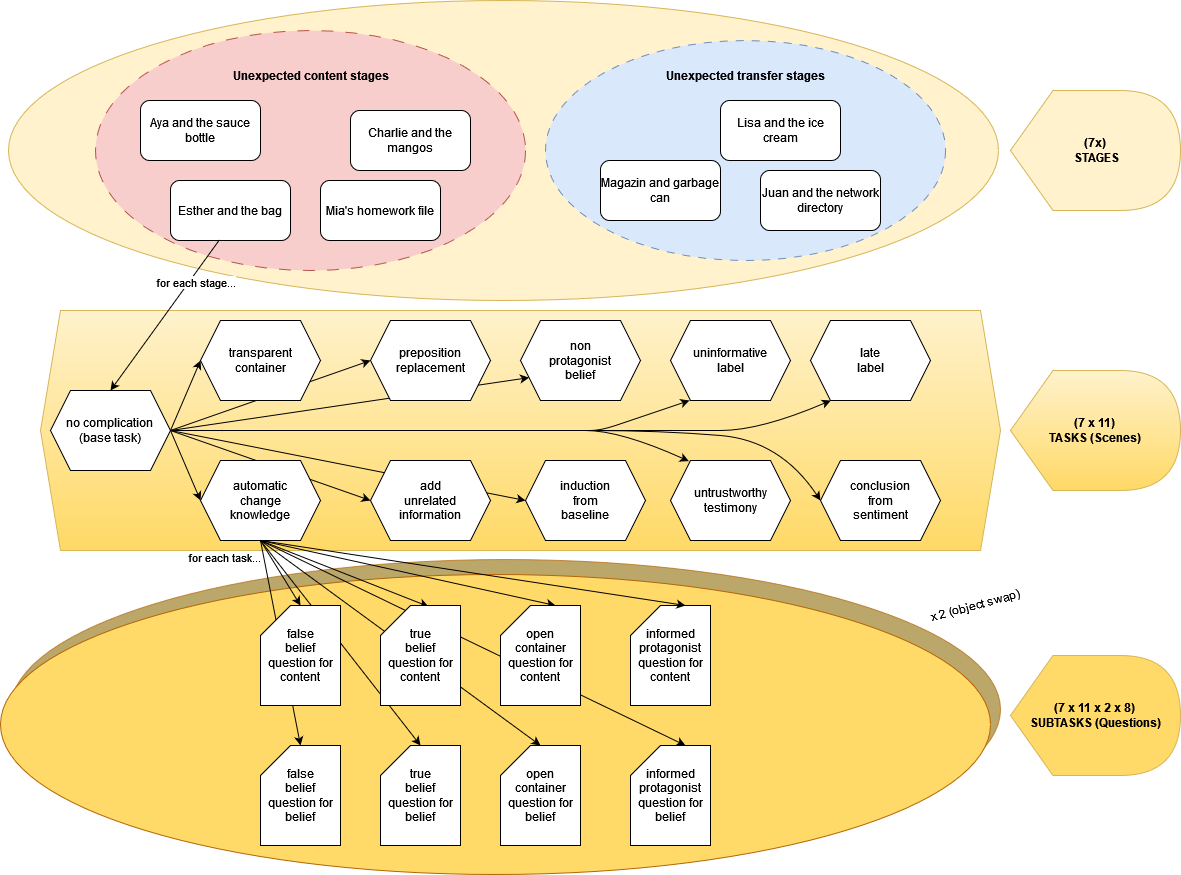}
	\caption[Structure of Our Dataset]{\textbf{Structure of our dataset.} Each of the 7 stages contains one unperturbed base task and up to 10 perturbed variants. Each task includes 16 subtasks probing different types of understanding.}
	\label{fig:dataset-structure}
\end{figure*}

The dataset structure from stage via task to subtasks can be observed in Figure \ref{fig:dataset-structure}

\subsection{Evaluation and Analysis Pipeline}

\begin{figure*}[h]
    \centering
    \includegraphics[width=\textwidth]{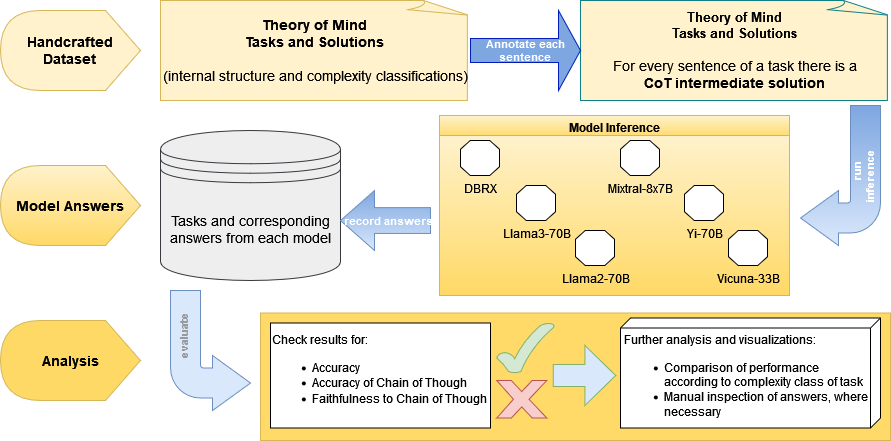}
    \caption[ToM Evaluation and Analysis Pipeline]{\textbf{ToM evaluation and analysis pipeline.} Our dataset is used to benchmark six open-source LLMs under Vanilla and Chain-of-Thought prompting. Model answers are evaluated for final answer accuracy, CoT correctness, and faithfulness.}
    \label{fig:experiments:pipe}
\end{figure*}

The whole process from dataset to evaluation results is illustrated in Figure \ref{fig:experiments:pipe}.
\label{sec:appendix:pipeline-figure}


%% file: tables/RQ3/accuracies.tex
\begin{landscape}
\begin{table}[p]
\centering

{\tiny

\setlength{\tabcolsep}{2pt}  

\begin{minipage}{\linewidth}

\begin{longtable*}{llllllllllllll}
\toprule
 \textbf{ALL SUBTASKS} & Model & \multicolumn{2}{l}{Llama-2-70b chat-hf} & \multicolumn{2}{l}{vicuna-33b v1.3} & \multicolumn{2}{l}{Mixtral-8x7B Instruct-v0.1} & \multicolumn{2}{l}{Yi-34B Chat} & \multicolumn{2}{l}{Meta-Llama-3-70B Instruct} & \multicolumn{2}{l}{dbrx-instruct} \\
 & PT & V-P & CoT-P & V-P & CoT-P & V-P & CoT-P & V-P & CoT-P & V-P & CoT-P & V-P & CoT-P \\
Perturbation Class & Accuracy &  &  &  &  &  &  &  &  &  &  &  &  \\
\midrule
\endfirsthead
\toprule
 & Model & \multicolumn{2}{l}{Llama-2-70b chat-hf} & \multicolumn{2}{l}{vicuna-33b v1.3} & \multicolumn{2}{l}{Mixtral-8x7B Instruct-v0.1} & \multicolumn{2}{l}{Yi-34B Chat} & \multicolumn{2}{l}{Meta-Llama-3-70B Instruct} & \multicolumn{2}{l}{dbrx-instruct} \\
 & PT & V-P & CoT-P & V-P & CoT-P & V-P & CoT-P & V-P & CoT-P & V-P & CoT-P & V-P & CoT-P \\
complexity & Accuracy &  &  &  &  &  &  &  &  &  &  &  &  \\
\midrule
\endhead
\midrule
\multicolumn{14}{r}{Continued on next page} \\
\midrule
\endfoot
\bottomrule
\endlastfoot
\multirow[t]{2}{*}{no complication} & meas. & 0.824 & 0.824 & 0.843 & 0.888 & 0.804 & 0.902 & 0.817 & 0.885 & 0.962 & 0.962 & 0.875 & 0.885 \\
 & $\text{CI}_{95\%}$ & [0.737, 0.886] & [0.737, 0.886] & [0.751, 0.905] & [0.803, 0.939] & [0.715, 0.870] & [0.826, 0.947] & [0.731, 0.880] & [0.807, 0.934] & [0.901, 0.988] & [0.901, 0.988] & [0.796, 0.926] & [0.807, 0.934] \\
\cline{1-14}
\multirow[t]{2}{*}{overall} & meas. & 0.783 & 0.764 & 0.805 & 0.79 & 0.715 & 0.784 & 0.78 & 0.759 & 0.853 & 0.878 & 0.776 & 0.773 \\
 & $\text{CI}_{95\%}$ & [0.757, 0.807] & [0.737, 0.789] & [0.778, 0.830] & [0.762, 0.815] & [0.687, 0.742] & [0.757, 0.808] & [0.753, 0.805] & [0.731, 0.785] & [0.830, 0.873] & [0.856, 0.896] & [0.749, 0.800] & [0.746, 0.797] \\
\cline{1-14}
\multirow[t]{2}{*}{transparent container} & meas. & 0.798 & 0.846 & 0.856 & 0.644 & 0.713 & 0.842 & 0.82 & 0.79 & 0.885 & 0.894 & 0.864 & 0.845 \\
 & $\text{CI}_{95\%}$ & [0.710, 0.864] & [0.763, 0.904] & [0.766, 0.914] & [0.541, 0.735] & [0.618, 0.792] & [0.756, 0.901] & [0.732, 0.883] & [0.699, 0.859] & [0.807, 0.934] & [0.818, 0.941] & [0.783, 0.918] & [0.761, 0.903] \\
\cline{1-14}
\multirow[t]{2}{*}{preposition replacement} & meas. & 0.7 & 0.688 & 0.809 & 0.691 & 0.474 & 0.641 & 0.75 & 0.75 & 0.772 & 0.823 & 0.7 & 0.743 \\
 & $\text{CI}_{95\%}$ & [0.592, 0.789] & [0.579, 0.778] & [0.698, 0.885] & [0.573, 0.788] & [0.368, 0.584] & [0.530, 0.738] & [0.641, 0.834] & [0.641, 0.834] & [0.667, 0.851] & [0.722, 0.892] & [0.584, 0.795] & [0.629, 0.831] \\
\cline{1-14}
\multirow[t]{2}{*}{uninformative label} & meas. & 0.703 & 0.594 & 0.75 & 0.817 & 0.812 & 0.766 & 0.688 & 0.719 & 0.8 & 0.833 & 0.75 & 0.828 \\
 & $\text{CI}_{95\%}$ & [0.581, 0.801] & [0.471, 0.705] & [0.626, 0.843] & [0.698, 0.895] & [0.698, 0.890] & [0.647, 0.853] & [0.565, 0.788] & [0.598, 0.814] & [0.680, 0.883] & [0.717, 0.908] & [0.630, 0.840] & [0.715, 0.902] \\
\cline{1-14}
\multirow[t]{2}{*}{late label} & meas. & 0.762 & 0.802 & 0.779 & 0.789 & 0.772 & 0.752 & 0.792 & 0.673 & 0.856 & 0.865 & 0.817 & 0.788 \\
 & $\text{CI}_{95\%}$ & [0.670, 0.835] & [0.713, 0.868] & [0.684, 0.851] & [0.696, 0.860] & [0.680, 0.843] & [0.659, 0.826] & [0.702, 0.860] & [0.576, 0.757] & [0.774, 0.911] & [0.785, 0.919] & [0.731, 0.880] & [0.699, 0.856] \\
\cline{1-14}
\multirow[t]{2}{*}{non protagonist belief} & meas. & 0.827 & 0.875 & 0.9 & 0.9 & 0.765 & 0.847 & 0.827 & 0.769 & 0.892 & 0.922 & 0.837 & 0.875 \\
 & $\text{CI}_{95\%}$ & [0.742, 0.888] & [0.796, 0.926] & [0.818, 0.948] & [0.818, 0.948] & [0.671, 0.838] & [0.761, 0.906] & [0.742, 0.888] & [0.679, 0.840] & [0.815, 0.940] & [0.850, 0.961] & [0.752, 0.896] & [0.796, 0.926] \\
\cline{1-14}
\multirow[t]{2}{*}{automatic change knowledge} & meas. & 0.7 & 0.625 & 0.656 & 0.672 & 0.588 & 0.55 & 0.608 & 0.49 & 0.756 & 0.718 & 0.625 & 0.512 \\
 & $\text{CI}_{95\%}$ & [0.592, 0.789] & [0.515, 0.723] & [0.533, 0.761] & [0.549, 0.774] & [0.478, 0.689] & [0.441, 0.654] & [0.471, 0.729] & [0.359, 0.623] & [0.649, 0.838] & [0.609, 0.806] & [0.515, 0.723] & [0.405, 0.619] \\
\cline{1-14}
\multirow[t]{2}{*}{add unrelated information} & meas. & 0.856 & 0.808 & 0.846 & 0.791 & 0.772 & 0.95 & 0.827 & 0.798 & 0.962 & 0.952 & 0.837 & 0.856 \\
 & $\text{CI}_{95\%}$ & [0.774, 0.911] & [0.720, 0.872] & [0.756, 0.907] & [0.696, 0.862] & [0.680, 0.843] & [0.886, 0.981] & [0.742, 0.888] & [0.710, 0.864] & [0.901, 0.988] & [0.889, 0.982] & [0.752, 0.896] & [0.774, 0.911] \\
\cline{1-14}
\multirow[t]{2}{*}{induction from baseline} & meas. & 0.851 & 0.759 & 0.829 & 0.878 & 0.693 & 0.807 & 0.798 & 0.821 & 0.841 & 0.92 & 0.716 & 0.705 \\
 & $\text{CI}_{95\%}$ & [0.759, 0.911] & [0.658, 0.837] & [0.732, 0.896] & [0.787, 0.934] & [0.590, 0.780] & [0.711, 0.876] & [0.698, 0.870] & [0.724, 0.889] & [0.749, 0.904] & [0.841, 0.963] & [0.613, 0.800] & [0.602, 0.790] \\
\cline{1-14}
\multirow[t]{2}{*}{untrustworthy testimony} & meas. & 0.729 & 0.76 & 0.733 & 0.779 & 0.719 & 0.802 & 0.74 & 0.781 & 0.777 & 0.904 & 0.76 & 0.812 \\
 & $\text{CI}_{95\%}$ & [0.632, 0.808] & [0.665, 0.835] & [0.630, 0.815] & [0.679, 0.854] & [0.621, 0.799] & [0.710, 0.870] & [0.643, 0.817] & [0.688, 0.852] & [0.681, 0.849] & [0.825, 0.950] & [0.665, 0.835] & [0.722, 0.878] \\
\cline{1-14}
\multirow[t]{2}{*}{conclusion from sentiment} & meas. & 0.804 & 0.706 & 0.797 & 0.797 & 0.709 & 0.68 & 0.788 & 0.731 & 0.808 & 0.798 & 0.673 & 0.596 \\
 & $\text{CI}_{95\%}$ & [0.715, 0.870] & [0.611, 0.786] & [0.694, 0.872] & [0.694, 0.872] & [0.614, 0.788] & [0.584, 0.762] & [0.699, 0.856] & [0.638, 0.807] & [0.720, 0.872] & [0.710, 0.864] & [0.578, 0.756] & [0.500, 0.685] \\
\cline{1-14}
\end{longtable*}


\begin{longtable*}{llllllllllllll}
\toprule
 \textbf{FALSE BELIEF SUBTASKS}& Model & \multicolumn{2}{l}{Llama-2-70b chat-hf} & \multicolumn{2}{l}{vicuna-33b v1.3} & \multicolumn{2}{l}{Mixtral-8x7B Instruct-v0.1} & \multicolumn{2}{l}{Yi-34B Chat} & \multicolumn{2}{l}{Meta-Llama-3-70B Instruct} & \multicolumn{2}{l}{dbrx-instruct} \\
 & PT & V-P & CoT-P & V-P & CoT-P & V-P & CoT-P & V-P & CoT-P & V-P & CoT-P & V-P & CoT-P \\
Perturbation Class & Accuracy &  &  &  &  &  &  &  &  &  &  &  &  \\
\midrule
\endfirsthead
\toprule
 & Model & \multicolumn{2}{r}{Llama-2-70b chat-hf} & \multicolumn{2}{r}{vicuna-33b v1.3} & \multicolumn{2}{r}{Mixtral-8x7B Instruct-v0.1} & \multicolumn{2}{r}{Yi-34B Chat} & \multicolumn{2}{r}{Meta-Llama-3-70B Instruct} & \multicolumn{2}{r}{dbrx-instruct} \\
 & PT & V-P & CoT-P & V-P & CoT-P & V-P & CoT-P & V-P & CoT-P & V-P & CoT-P & V-P & CoT-P \\
complexity & Accuracy &  &  &  &  &  &  &  &  &  &  &  &  \\
\midrule
\endhead
\midrule
\multicolumn{14}{r}{Continued on next page} \\
\midrule
\endfoot
\bottomrule
\endlastfoot
\multirow[t]{2}{*}{no complication} & meas. & 0.615 & 0.769 & 0.273 & 0.818 & 0.929 & 0.786 & 0.429 & 1.0 & 1 & 1.0 & 0.786 & 1.0 \\
 & $\text{CI}_{95\%}$ & [0.354, 0.822] & [0.489, 0.922] & [0.095, 0.572] & [0.510, 0.957] & [0.661, 1.000] & [0.515, 0.929] & [0.215, 0.674] & [0.744, 1.000] & [0.744, 1.000] & [0.744, 1.000] & [0.515, 0.929] & [0.744, 1.000] \\
\cline{1-14}
\multirow[t]{2}{*}{overall} & meas. & 0.466 & 0.466 & 0.427 & 0.496 & 0.474 & 0.504 & 0.465 & 0.488 & 0.622 & 0.704 & 0.537 & 0.575 \\
 & $\text{CI}_{95\%}$ & [0.384, 0.551] & [0.384, 0.551] & [0.342, 0.518] & [0.407, 0.585] & [0.392, 0.558] & [0.420, 0.587] & [0.381, 0.551] & [0.404, 0.574] & [0.538, 0.699] & [0.622, 0.774] & [0.453, 0.619] & [0.490, 0.655] \\
\cline{1-14}
\multirow[t]{2}{*}{transparent container} & meas. & 0.286 & 0.286 & 0.692 & 0.538 & 0.071 & 0.357 & 0.429 & 0.286 & 0.429 & 0.571 & 0.429 & 0.571 \\
 & $\text{CI}_{95\%}$ & [0.116, 0.551] & [0.116, 0.551] & [0.420, 0.874] & [0.292, 0.767] & [0.000, 0.339] & [0.164, 0.614] & [0.215, 0.674] & [0.116, 0.551] & [0.215, 0.674] & [0.326, 0.785] & [0.215, 0.674] & [0.326, 0.785] \\
\cline{1-14}
\multirow[t]{2}{*}{preposition replacement} & meas. & 0.2 & 0 & 0.4 & 0.3 & 0.2 & 0.2 & 0.6 & 0.3 & 0.2 & 0.4 & 0.375 & 0.25 \\
 & $\text{CI}_{95\%}$ & [0.049, 0.522] & [0.000, 0.326] & [0.169, 0.688] & [0.106, 0.608] & [0.049, 0.522] & [0.049, 0.522] & [0.312, 0.831] & [0.106, 0.608] & [0.049, 0.522] & [0.169, 0.688] & [0.138, 0.696] & [0.067, 0.600] \\
\cline{1-14}
\multirow[t]{2}{*}{uninformative label} & meas. & 0.125 & 0 & 0.375 & 0.25 & 0.5 & 0.25 & 0.25 & 0.125 & 0.25 & 0.375 & 0.625 & 0.625 \\
 & $\text{CI}_{95\%}$ & [0.005, 0.495] & [0.000, 0.378] & [0.138, 0.696] & [0.067, 0.600] & [0.217, 0.783] & [0.067, 0.600] & [0.067, 0.600] & [0.005, 0.495] & [0.067, 0.600] & [0.138, 0.696] & [0.304, 0.862] & [0.304, 0.862] \\
\cline{1-14}
\multirow[t]{2}{*}{late label} & meas. & 0.538 & 0.615 & 0.333 & 0.75 & 0.429 & 0.429 & 0.385 & 0.231 & 0.429 & 0.429 & 0.643 & 0.571 \\
 & $\text{CI}_{95\%}$ & [0.292, 0.767] & [0.354, 0.822] & [0.138, 0.612] & [0.460, 0.915] & [0.215, 0.674] & [0.215, 0.674] & [0.178, 0.646] & [0.078, 0.511] & [0.215, 0.674] & [0.215, 0.674] & [0.386, 0.836] & [0.326, 0.785] \\
\cline{1-14}
\multirow[t]{2}{*}{non protagonist belief} & meas. & 0.643 & 0.714 & 0.667 & 0.778 & 0.692 & 0.846 & 0.571 & 0.5 & 0.923 & 1.0 & 0.786 & 0.857 \\
 & $\text{CI}_{95\%}$ & [0.386, 0.836] & [0.449, 0.884] & [0.351, 0.880] & [0.441, 0.943] & [0.420, 0.874] & [0.563, 0.966] & [0.326, 0.785] & [0.269, 0.731] & [0.642, 1.000] & [0.729, 1.000] & [0.515, 0.929] & [0.586, 0.970] \\
\cline{1-14}
\multirow[t]{2}{*}{automatic change knowledge} & meas. & 0.6 & 0.4 & 0.375 & 0.25 & 0.6 & 0.3 & 0.667 & 0.667 & 0.5 & 0.5 & 0.6 & 0.5 \\
 & $\text{CI}_{95\%}$ & [0.312, 0.831] & [0.169, 0.688] & [0.138, 0.696] & [0.067, 0.600] & [0.312, 0.831] & [0.106, 0.608] & [0.296, 0.904] & [0.296, 0.904] & [0.238, 0.762] & [0.238, 0.762] & [0.312, 0.831] & [0.238, 0.762] \\
\cline{1-14}
\multirow[t]{2}{*}{add unrelated information} & meas. & 0.571 & 0.857 & 0.5 & 0.357 & 0.643 & 1.0 & 0.643 & 0.643 & 1 & 1.0 & 0.643 & 0.857 \\
 & $\text{CI}_{95\%}$ & [0.326, 0.785] & [0.586, 0.970] & [0.269, 0.731] & [0.164, 0.614] & [0.386, 0.836] & [0.744, 1.000] & [0.386, 0.836] & [0.386, 0.836] & [0.744, 1.000] & [0.744, 1.000] & [0.386, 0.836] & [0.586, 0.970] \\
\cline{1-14}
\multirow[t]{2}{*}{induction from baseline} & meas. & 0.545 & 0.364 & 0.636 & 0.455 & 0.5 & 0.5 & 0.364 & 0.545 & 0.667 & 0.917 & 0.333 & 0.333 \\
 & $\text{CI}_{95\%}$ & [0.281, 0.786] & [0.152, 0.648] & [0.352, 0.848] & [0.214, 0.719] & [0.255, 0.745] & [0.255, 0.745] & [0.152, 0.648] & [0.281, 0.786] & [0.388, 0.862] & [0.621, 1.000] & [0.138, 0.612] & [0.138, 0.612] \\
\cline{1-14}
\multirow[t]{2}{*}{untrustworthy testimony} & meas. & 0.333 & 0.583 & 0 & 0.5 & 0.5 & 0.667 & 0.333 & 0.667 & 0.667 & 0.75 & 0.333 & 0.5 \\
 & $\text{CI}_{95\%}$ & [0.138, 0.612] & [0.319, 0.806] & [0.000, 0.326] & [0.238, 0.762] & [0.255, 0.745] & [0.388, 0.862] & [0.138, 0.612] & [0.388, 0.862] & [0.388, 0.862] & [0.460, 0.915] & [0.138, 0.612] & [0.255, 0.745] \\
\cline{1-14}
\multirow[t]{2}{*}{conclusion from sentiment} & meas. & 0.5 & 0.214 & 0.364 & 0.364 & 0.143 & 0 & 0.462 & 0.308 & 0.5 & 0.571 & 0.286 & 0.071 \\
 & $\text{CI}_{95\%}$ & [0.269, 0.731] & [0.071, 0.485] & [0.152, 0.648] & [0.152, 0.648] & [0.030, 0.414] & [0.000, 0.256] & [0.233, 0.708] & [0.126, 0.580] & [0.269, 0.731] & [0.326, 0.785] & [0.116, 0.551] & [0.000, 0.339] \\
\cline{1-14}
\end{longtable*}


\end{minipage}
}

\caption[Theory of Mind Accuracies: All Models and Perturbation Classes]{Theory of Mind Accuracies: All Models and Perturbation Classes. Upper table created using "All-Subtasks". Lower table measurements on "False Belief" Tasks. V-P: Vanilla - Prompting; CoT-P: Chain of Thought - Prompting.}
\label{tab:RQ3:accuracies}
\end{table}
\end{landscape}

%% file: tables/RQ3/effect_strengths_pert_treatment_all_subtasks.tex
\begin{landscape}
\begin{table}[p]
\centering

{\tiny

\setlength{\tabcolsep}{1pt}  

\begin{minipage}{\linewidth}

\begin{longtable*}{lllllllllllllll}
\toprule
 &  & Model & \multicolumn{2}{l}{meta-llama/Llama-2-70b-chat-hf} & \multicolumn{2}{l}{lmsys/vicuna-33b-v1.3} & \multicolumn{2}{l}{mistralai/Mixtral-8x7B-Instruct-v0.1} & \multicolumn{2}{l}{01-ai/Yi-34B-Chat} & \multicolumn{2}{l}{meta-llama/Meta-Llama-3-70B-Instruct} & \multicolumn{2}{l}{databricks/dbrx-instruct} \\
 &  & Effect & ATE & RR & ATE & RR & ATE & RR & ATE & RR & ATE & RR & ATE & RR \\
CoT & Perturbation Class & Statistic &  &  &  &  &  &  &  &  &  &  &  &  \\
\midrule
\endfirsthead
\toprule
 &  & Model & \multicolumn{2}{l}{meta-llama/Llama-2-70b-chat-hf} & \multicolumn{2}{l}{lmsys/vicuna-33b-v1.3} & \multicolumn{2}{l}{mistralai/Mixtral-8x7B-Instruct-v0.1} & \multicolumn{2}{l}{01-ai/Yi-34B-Chat} & \multicolumn{2}{l}{meta-llama/Meta-Llama-3-70B-Instruct} & \multicolumn{2}{l}{databricks/dbrx-instruct} \\
 &  & Effect & ATE & RR & ATE & RR & ATE & RR & ATE & RR & ATE & RR & ATE & RR \\
CoT & Perturbation Class & Statistic &  &  &  &  &  &  &  &  &  &  &  &  \\
\midrule
\endhead
\midrule
\multicolumn{15}{r}{Continued on next page} \\
\midrule
\endfoot
\bottomrule
\endlastfoot
\multirow[t]{22}{*}{Without CoT} & \multirow[t]{2}{*}{Overall} & meas. & -0.040 & 0.951 & -0.037 & 0.956 & -0.089 & 0.890 & -0.037 & 0.955 & -0.108 & 0.887 & -0.099 & 0.887 \\
 &  & $\text{CI}_{95\%}$ & [-0.119, 0.038] & [0.864, 1.047] & [-0.118, 0.044] & [0.868, 1.053] & [-0.171, -0.006] & [0.802, 0.987] & [-0.116, 0.042] & [0.866, 1.052] & [-0.157, -0.060] & [0.843, 0.934] & [-0.169, -0.029] & [0.817, 0.962] \\
\cline{2-15}
 & \multirow[t]{2}{*}{Transparent Container} & meas. & -0.025 & 0.969 & 0.013 & 1.015 & -0.091 & 0.887 & 0.003 & 1.003 & -0.077 & 0.920 & -0.011 & 0.988 \\
 &  & $\text{CI}_{95\%}$ & [-0.133, 0.082] & [0.849, 1.106] & [-0.094, 0.119] & [0.895, 1.151] & [-0.208, 0.025] & [0.759, 1.036] & [-0.104, 0.109] & [0.881, 1.142] & [-0.154, -0.000] & [0.845, 1.001] & [-0.105, 0.083] & [0.886, 1.100] \\
\cline{2-15}
 & \multirow[t]{2}{*}{Preposition Replacement} & meas. & -0.124 & 0.850 & -0.034 & 0.960 & -0.330 & 0.590 & -0.067 & 0.918 & -0.189 & 0.803 & -0.175 & 0.800 \\
 &  & $\text{CI}_{95\%}$ & [-0.247, 0.000] & [0.719, 1.005] & [-0.155, 0.087] & [0.828, 1.112] & [-0.462, -0.197] & [0.461, 0.756] & [-0.189, 0.055] & [0.784, 1.074] & [-0.291, -0.088] & [0.707, 0.912] & [-0.299, -0.051] & [0.676, 0.946] \\
\cline{2-15}
 & \multirow[t]{2}{*}{Uninformative Label} & meas. & -0.120 & 0.854 & -0.093 & 0.890 & 0.009 & 1.011 & -0.130 & 0.841 & -0.162 & 0.832 & -0.125 & 0.857 \\
 &  & $\text{CI}_{95\%}$ & [-0.253, 0.012] & [0.713, 1.023] & [-0.225, 0.040] & [0.750, 1.056] & [-0.115, 0.132] & [0.868, 1.177] & [-0.264, 0.004] & [0.699, 1.013] & [-0.272, -0.051] & [0.727, 0.952] & [-0.248, -0.002] & [0.732, 1.004] \\
\cline{2-15}
 & \multirow[t]{2}{*}{Late Label} & meas. & -0.061 & 0.926 & -0.064 & 0.924 & -0.032 & 0.961 & -0.025 & 0.969 & -0.106 & 0.890 & -0.058 & 0.934 \\
 &  & $\text{CI}_{95\%}$ & [-0.172, 0.050] & [0.804, 1.066] & [-0.177, 0.049] & [0.803, 1.064] & [-0.144, 0.081] & [0.833, 1.108] & [-0.134, 0.084] & [0.846, 1.110] & [-0.187, -0.025] & [0.812, 0.976] & [-0.157, 0.041] & [0.830, 1.051] \\
\cline{2-15}
 & \multirow[t]{2}{*}{Non Protagonist Belief} & meas. & 0.003 & 1.004 & 0.057 & 1.068 & -0.039 & 0.952 & 0.010 & 1.012 & -0.069 & 0.928 & -0.038 & 0.956 \\
 &  & $\text{CI}_{95\%}$ & [-0.101, 0.108] & [0.885, 1.140] & [-0.043, 0.158] & [0.951, 1.200] & [-0.152, 0.075] & [0.823, 1.101] & [-0.095, 0.114] & [0.891, 1.149] & [-0.145, 0.007] & [0.854, 1.008] & [-0.135, 0.059] & [0.853, 1.071] \\
\cline{2-15}
 & \multirow[t]{2}{*}{Automatic Change Knowledge} & meas. & -0.124 & 0.850 & -0.186 & 0.779 & -0.216 & 0.731 & -0.209 & 0.744 & -0.205 & 0.787 & -0.250 & 0.714 \\
 &  & $\text{CI}_{95\%}$ & [-0.247, 0.000] & [0.719, 1.005] & [-0.324, -0.049] & [0.640, 0.947] & [-0.347, -0.086] & [0.596, 0.896] & [-0.359, -0.060] & [0.590, 0.938] & [-0.309, -0.101] & [0.689, 0.898] & [-0.373, -0.127] & [0.595, 0.857] \\
\cline{2-15}
 & \multirow[t]{2}{*}{Add Unrelated Information} & meas. & 0.032 & 1.039 & 0.003 & 1.004 & -0.032 & 0.961 & 0.010 & 1.012 & 0.000 & 1.000 & -0.038 & 0.956 \\
 &  & $\text{CI}_{95\%}$ & [-0.069, 0.134] & [0.921, 1.173] & [-0.104, 0.111] & [0.884, 1.140] & [-0.144, 0.081] & [0.833, 1.108] & [-0.095, 0.114] & [0.891, 1.149] & [-0.061, 0.061] & [0.938, 1.066] & [-0.135, 0.059] & [0.853, 1.071] \\
\cline{2-15}
 & \multirow[t]{2}{*}{Induction From Baseline} & meas. & 0.027 & 1.033 & -0.013 & 0.984 & -0.111 & 0.862 & -0.020 & 0.976 & -0.121 & 0.875 & -0.159 & 0.818 \\
 &  & $\text{CI}_{95\%}$ & [-0.080, 0.134] & [0.909, 1.173] & [-0.126, 0.099] & [0.860, 1.126] & [-0.233, 0.012] & [0.729, 1.019] & [-0.134, 0.094] & [0.847, 1.124] & [-0.209, -0.032] & [0.789, 0.969] & [-0.273, -0.045] & [0.704, 0.950] \\
\cline{2-15}
 & \multirow[t]{2}{*}{Untrustworthy Testimony} & meas. & -0.094 & 0.885 & -0.110 & 0.869 & -0.085 & 0.894 & -0.078 & 0.905 & -0.185 & 0.808 & -0.115 & 0.869 \\
 &  & $\text{CI}_{95\%}$ & [-0.210, 0.021] & [0.761, 1.030] & [-0.230, 0.010] & [0.744, 1.016] & [-0.203, 0.033] & [0.764, 1.046] & [-0.192, 0.037] & [0.780, 1.050] & [-0.279, -0.091] & [0.718, 0.908] & [-0.222, -0.008] & [0.760, 0.994] \\
\cline{2-15}
 & \multirow[t]{2}{*}{Conclusion From Sentiment} & meas. & -0.020 & 0.976 & -0.045 & 0.946 & -0.095 & 0.882 & -0.029 & 0.965 & -0.154 & 0.840 & -0.202 & 0.769 \\
 &  & $\text{CI}_{95\%}$ & [-0.127, 0.088] & [0.856, 1.114] & [-0.162, 0.072] & [0.820, 1.093] & [-0.211, 0.021] & [0.755, 1.030] & [-0.137, 0.079] & [0.843, 1.104] & [-0.241, -0.066] & [0.757, 0.932] & [-0.312, -0.092] & [0.661, 0.895] \\
\cline{1-15} \cline{2-15}
\multirow[t]{22}{*}{With CoT} & \multirow[t]{2}{*}{Overall} & meas. & -0.060 & 0.927 & -0.098 & 0.890 & -0.118 & 0.869 & -0.126 & 0.858 & -0.084 & 0.913 & -0.112 & 0.874 \\
 &  & $\text{CI}_{95\%}$ & [-0.139, 0.019] & [0.842, 1.021] & [-0.171, -0.025] & [0.818, 0.968] & [-0.184, -0.053] & [0.807, 0.936] & [-0.194, -0.057] & [0.792, 0.929] & [-0.132, -0.036] & [0.868, 0.960] & [-0.180, -0.043] & [0.807, 0.945] \\
\cline{2-15}
 & \multirow[t]{2}{*}{Transparent Container} & meas. & 0.023 & 1.027 & -0.243 & 0.726 & -0.060 & 0.933 & -0.095 & 0.893 & -0.067 & 0.930 & -0.040 & 0.955 \\
 &  & $\text{CI}_{95\%}$ & [-0.080, 0.125] & [0.909, 1.162] & [-0.362, -0.125] & [0.613, 0.860] & [-0.154, 0.034] & [0.837, 1.040] & [-0.196, 0.007] & [0.789, 1.011] & [-0.142, 0.008] & [0.857, 1.010] & [-0.135, 0.055] & [0.855, 1.066] \\
\cline{2-15}
 & \multirow[t]{2}{*}{Preposition Replacement} & meas. & -0.136 & 0.835 & -0.196 & 0.779 & -0.261 & 0.711 & -0.135 & 0.848 & -0.139 & 0.856 & -0.142 & 0.840 \\
 &  & $\text{CI}_{95\%}$ & [-0.261, -0.011] & [0.704, 0.991] & [-0.324, -0.069] & [0.655, 0.926] & [-0.381, -0.140] & [0.596, 0.847] & [-0.250, -0.019] & [0.732, 0.982] & [-0.234, -0.044] & [0.765, 0.958] & [-0.261, -0.022] & [0.720, 0.979] \\
\cline{2-15}
 & \multirow[t]{2}{*}{Uninformative Label} & meas. & -0.230 & 0.721 & -0.071 & 0.920 & -0.136 & 0.849 & -0.166 & 0.812 & -0.128 & 0.867 & -0.056 & 0.936 \\
 &  & $\text{CI}_{95\%}$ & [-0.368, -0.091] & [0.580, 0.896] & [-0.191, 0.049] & [0.797, 1.062] & [-0.256, -0.017] & [0.731, 0.986] & [-0.291, -0.040] & [0.688, 0.960] & [-0.233, -0.023] & [0.766, 0.980] & [-0.169, 0.056] & [0.819, 1.070] \\
\cline{2-15}
 & \multirow[t]{2}{*}{Late Label} & meas. & -0.022 & 0.974 & -0.098 & 0.889 & -0.149 & 0.834 & -0.211 & 0.761 & -0.096 & 0.900 & -0.096 & 0.891 \\
 &  & $\text{CI}_{95\%}$ & [-0.129, 0.086] & [0.853, 1.112] & [-0.205, 0.008] & [0.782, 1.012] & [-0.253, -0.046] & [0.733, 0.950] & [-0.322, -0.101] & [0.654, 0.886] & [-0.176, -0.016] & [0.823, 0.984] & [-0.197, 0.005] & [0.789, 1.008] \\
\cline{2-15}
 & \multirow[t]{2}{*}{Non Protagonist Belief} & meas. & 0.051 & 1.062 & 0.012 & 1.014 & -0.055 & 0.939 & -0.115 & 0.870 & -0.040 & 0.958 & -0.010 & 0.989 \\
 &  & $\text{CI}_{95\%}$ & [-0.048, 0.150] & [0.945, 1.195] & [-0.082, 0.107] & [0.913, 1.127] & [-0.149, 0.039] & [0.842, 1.047] & [-0.218, -0.013] & [0.766, 0.987] & [-0.110, 0.030] & [0.889, 1.033] & [-0.101, 0.081] & [0.892, 1.097] \\
\cline{2-15}
 & \multirow[t]{2}{*}{Automatic Change Knowledge} & meas. & -0.199 & 0.759 & -0.216 & 0.757 & -0.352 & 0.610 & -0.394 & 0.554 & -0.244 & 0.747 & -0.372 & 0.579 \\
 &  & $\text{CI}_{95\%}$ & [-0.326, -0.071] & [0.628, 0.917] & [-0.347, -0.084] & [0.630, 0.910] & [-0.474, -0.230] & [0.497, 0.748] & [-0.541, -0.248] & [0.419, 0.732] & [-0.351, -0.136] & [0.646, 0.863] & [-0.496, -0.248] & [0.465, 0.722] \\
\cline{2-15}
 & \multirow[t]{2}{*}{Add Unrelated Information} & meas. & -0.016 & 0.981 & -0.096 & 0.891 & 0.049 & 1.054 & -0.087 & 0.902 & -0.010 & 0.990 & -0.029 & 0.967 \\
 &  & $\text{CI}_{95\%}$ & [-0.122, 0.091] & [0.861, 1.117] & [-0.204, 0.011] & [0.782, 1.016] & [-0.028, 0.125] & [0.969, 1.146] & [-0.186, 0.013] & [0.800, 1.018] & [-0.073, 0.054] & [0.926, 1.058] & [-0.122, 0.065] & [0.869, 1.077] \\
\cline{2-15}
 & \multirow[t]{2}{*}{Induction From Baseline} & meas. & -0.065 & 0.921 & -0.010 & 0.989 & -0.095 & 0.895 & -0.063 & 0.929 & -0.041 & 0.957 & -0.180 & 0.796 \\
 &  & $\text{CI}_{95\%}$ & [-0.181, 0.051] & [0.794, 1.069] & [-0.110, 0.090] & [0.883, 1.108] & [-0.198, 0.007] & [0.791, 1.011] & [-0.167, 0.041] & [0.821, 1.050] & [-0.116, 0.033] & [0.884, 1.037] & [-0.293, -0.067] & [0.685, 0.927] \\
\cline{2-15}
 & \multirow[t]{2}{*}{Untrustworthy Testimony} & meas. & -0.063 & 0.923 & -0.109 & 0.878 & -0.100 & 0.889 & -0.103 & 0.883 & -0.057 & 0.940 & -0.072 & 0.918 \\
 &  & $\text{CI}_{95\%}$ & [-0.176, 0.050] & [0.800, 1.066] & [-0.219, 0.002] & [0.766, 1.005] & [-0.200, 0.000] & [0.789, 1.003] & [-0.207, 0.001] & [0.777, 1.003] & [-0.133, 0.019] & [0.866, 1.021] & [-0.173, 0.029] & [0.814, 1.036] \\
\cline{2-15}
 & \multirow[t]{2}{*}{Conclusion From Sentiment} & meas. & -0.118 & 0.857 & -0.090 & 0.898 & -0.222 & 0.753 & -0.154 & 0.826 & -0.163 & 0.830 & -0.288 & 0.674 \\
 &  & $\text{CI}_{95\%}$ & [-0.232, -0.003] & [0.735, 0.999] & [-0.202, 0.022] & [0.785, 1.028] & [-0.330, -0.115] & [0.651, 0.873] & [-0.259, -0.048] & [0.721, 0.946] & [-0.252, -0.075] & [0.746, 0.923] & [-0.401, -0.176] & [0.568, 0.800] \\
\cline{1-15} \cline{2-15}
\end{longtable*}



\end{minipage}

}

\caption[Perturbation Effect Strengths: All Models. All-Subtasks.]{Perturbation Effect Strengths: All Models. All-Subtasks. ATE: Average Treatment Effect; RR: Relative Risk.}

\label{tab:RQ3:effect_strengths_pert_treatment_all_subtasks}
\end{table}
\end{landscape}

%% file: tables/RQ3/effect_strengths_pert_treatment_false_belief.tex
\begin{landscape}
\begin{table}[p]
\centering

{\tiny

\setlength{\tabcolsep}{1pt}  

\begin{minipage}{\linewidth}

\begin{longtable*}{lllllllllllllll}
\toprule
 &  & Model & \multicolumn{2}{l}{meta-llama/Llama-2-70b-chat-hf} & \multicolumn{2}{l}{lmsys/vicuna-33b-v1.3} & \multicolumn{2}{l}{mistralai/Mixtral-8x7B-Instruct-v0.1} & \multicolumn{2}{l}{01-ai/Yi-34B-Chat} & \multicolumn{2}{l}{meta-llama/Meta-Llama-3-70B-Instruct} & \multicolumn{2}{l}{databricks/dbrx-instruct} \\
 &  & Effect & ATE & RR & ATE & RR & ATE & RR & ATE & RR & ATE & RR & ATE & RR \\
CoT & Perturbation Class & Statistic &  &  &  &  &  &  &  &  &  &  &  &  \\
\midrule
\endfirsthead
\toprule
 &  & Model & \multicolumn{2}{l}{meta-llama/Llama-2-70b-chat-hf} & \multicolumn{2}{l}{lmsys/vicuna-33b-v1.3} & \multicolumn{2}{l}{mistralai/Mixtral-8x7B-Instruct-v0.1} & \multicolumn{2}{l}{01-ai/Yi-34B-Chat} & \multicolumn{2}{l}{meta-llama/Meta-Llama-3-70B-Instruct} & \multicolumn{2}{l}{databricks/dbrx-instruct} \\
 &  & Effect & ATE & RR & ATE & RR & ATE & RR & ATE & RR & ATE & RR & ATE & RR \\
CoT & Perturbation Class & Statistic &  &  &  &  &  &  &  &  &  &  &  &  \\
\midrule
\endhead
\midrule
\multicolumn{15}{r}{Continued on next page} \\
\midrule
\endfoot
\bottomrule
\endlastfoot
\multirow[t]{22}{*}{Without CoT} & \multirow[t]{2}{*}{Overall} & meas. & -0.149 & 0.758 & 0.155 & 1.567 & -0.454 & 0.511 & 0.037 & 1.085 & -0.378 & 0.622 & -0.248 & 0.684 \\
 &  & $\text{CI}_{95\%}$ & [-0.398, 0.099] & [0.498, 1.153] & [-0.100, 0.409] & [0.638, 3.849] & [-0.643, -0.266] & [0.396, 0.657] & [-0.208, 0.281] & [0.616, 1.911] & [-0.529, -0.226] & [0.518, 0.747] & [-0.471, -0.025] & [0.504, 0.928] \\
\cline{2-15}
 & \multirow[t]{2}{*}{Transparent Container} & meas. & -0.330 & 0.464 & 0.420 & 2.538 & -0.857 & 0.077 & 0.000 & 1.000 & -0.571 & 0.429 & -0.357 & 0.545 \\
 &  & $\text{CI}_{95\%}$ & [-0.649, -0.010] & [0.198, 1.088] & [0.090, 0.749] & [0.997, 6.461] & [-1.097, -0.618] & [0.007, 0.830] & [-0.325, 0.325] & [0.469, 2.133] & [-0.834, -0.309] & [0.247, 0.743] & [-0.666, -0.048] & [0.300, 0.991] \\
\cline{2-15}
 & \multirow[t]{2}{*}{Preposition Replacement} & meas. & -0.415 & 0.325 & 0.127 & 1.467 & -0.729 & 0.215 & 0.171 & 1.400 & -0.800 & 0.200 & -0.411 & 0.477 \\
 &  & $\text{CI}_{95\%}$ & [-0.748, -0.083] & [0.094, 1.126] & [-0.225, 0.480] & [0.494, 4.356] & [-1.020, -0.438] & [0.065, 0.713] & [-0.175, 0.518] & [0.703, 2.786] & [-1.069, -0.531] & [0.061, 0.658] & [-0.758, -0.063] & [0.217, 1.051] \\
\cline{2-15}
 & \multirow[t]{2}{*}{Uninformative Label} & meas. & -0.490 & 0.203 & 0.102 & 1.375 & -0.429 & 0.538 & -0.179 & 0.583 & -0.750 & 0.250 & -0.161 & 0.795 \\
 &  & $\text{CI}_{95\%}$ & [-0.829, -0.152] & [0.028, 1.496] & [-0.265, 0.469] & [0.436, 4.335] & [-0.758, -0.099] & [0.297, 0.976] & [-0.530, 0.173] & [0.177, 1.925] & [-1.046, -0.454] & [0.085, 0.732] & [-0.508, 0.187] & [0.474, 1.336] \\
\cline{2-15}
 & \multirow[t]{2}{*}{Late Label} & meas. & -0.077 & 0.875 & 0.061 & 1.222 & -0.500 & 0.462 & -0.044 & 0.897 & -0.571 & 0.429 & -0.143 & 0.818 \\
 &  & $\text{CI}_{95\%}$ & [-0.410, 0.256] & [0.489, 1.566] & [-0.276, 0.397] & [0.396, 3.775] & [-0.785, -0.215] & [0.262, 0.813] & [-0.372, 0.284] & [0.399, 2.018] & [-0.834, -0.309] & [0.247, 0.743] & [-0.449, 0.163] & [0.528, 1.268] \\
\cline{2-15}
 & \multirow[t]{2}{*}{Non Protagonist Belief} & meas. & 0.027 & 1.045 & 0.394 & 2.444 & -0.236 & 0.746 & 0.143 & 1.333 & -0.077 & 0.923 & 0.000 & 1.000 \\
 &  & $\text{CI}_{95\%}$ & [-0.297, 0.352] & [0.623, 1.752] & [0.038, 0.750] & [0.936, 6.387] & [-0.520, 0.047] & [0.512, 1.085] & [-0.182, 0.468] & [0.683, 2.604] & [-0.297, 0.143] & [0.732, 1.164] & [-0.293, 0.293] & [0.689, 1.451] \\
\cline{2-15}
 & \multirow[t]{2}{*}{Automatic Change Knowledge} & meas. & -0.015 & 0.975 & 0.102 & 1.375 & -0.329 & 0.646 & 0.238 & 1.556 & -0.500 & 0.500 & -0.186 & 0.764 \\
 &  & $\text{CI}_{95\%}$ & [-0.365, 0.334] & [0.548, 1.734] & [-0.265, 0.469] & [0.436, 4.335] & [-0.638, -0.019] & [0.404, 1.033] & [-0.143, 0.619] & [0.770, 3.142] & [-0.792, -0.208] & [0.292, 0.857] & [-0.517, 0.146] & [0.460, 1.267] \\
\cline{2-15}
 & \multirow[t]{2}{*}{Add Unrelated Information} & meas. & -0.044 & 0.929 & 0.227 & 1.833 & -0.286 & 0.692 & 0.214 & 1.500 & 0.000 & 1.000 & -0.143 & 0.818 \\
 &  & $\text{CI}_{95\%}$ & [-0.372, 0.284] & [0.534, 1.614] & [-0.105, 0.559] & [0.682, 4.930] & [-0.568, -0.004] & [0.466, 1.028] & [-0.107, 0.536] & [0.791, 2.845] & [-0.181, 0.181] & [0.834, 1.199] & [-0.449, 0.163] & [0.528, 1.268] \\
\cline{2-15}
 & \multirow[t]{2}{*}{Induction From Baseline} & meas. & -0.070 & 0.886 & 0.364 & 2.333 & -0.429 & 0.538 & -0.065 & 0.848 & -0.333 & 0.667 & -0.452 & 0.424 \\
 &  & $\text{CI}_{95\%}$ & [-0.414, 0.274] & [0.487, 1.613] & [0.020, 0.708] & [0.896, 6.079] & [-0.726, -0.131] & [0.319, 0.908] & [-0.403, 0.273] & [0.357, 2.019] & [-0.603, -0.064] & [0.457, 0.973] & [-0.767, -0.138] & [0.199, 0.906] \\
\cline{2-15}
 & \multirow[t]{2}{*}{Untrustworthy Testimony} & meas. & -0.282 & 0.542 & -0.273 & 0.000 & -0.429 & 0.538 & -0.095 & 0.778 & -0.333 & 0.667 & -0.452 & 0.424 \\
 &  & $\text{CI}_{95\%}$ & [-0.615, 0.051] & [0.242, 1.214] & [-0.562, 0.016] & [0.000, nan] & [-0.726, -0.131] & [0.319, 0.908] & [-0.425, 0.235] & [0.319, 1.895] & [-0.603, -0.064] & [0.457, 0.973] & [-0.767, -0.138] & [0.199, 0.906] \\
\cline{2-15}
 & \multirow[t]{2}{*}{Conclusion From Sentiment} & meas. & -0.115 & 0.812 & 0.091 & 1.333 & -0.786 & 0.154 & 0.033 & 1.077 & -0.500 & 0.500 & -0.500 & 0.364 \\
 &  & $\text{CI}_{95\%}$ & [-0.444, 0.213] & [0.447, 1.478] & [-0.253, 0.435] & [0.440, 4.042] & [-1.042, -0.530] & [0.040, 0.597] & [-0.297, 0.363] & [0.513, 2.263] & [-0.764, -0.236] & [0.310, 0.808] & [-0.800, -0.200] & [0.162, 0.815] \\
\cline{1-15} \cline{2-15}
\multirow[t]{22}{*}{With CoT} & \multirow[t]{2}{*}{Overall} & meas. & -0.303 & 0.606 & -0.322 & 0.606 & -0.282 & 0.641 & -0.512 & 0.488 & -0.296 & 0.704 & -0.425 & 0.575 \\
 &  & $\text{CI}_{95\%}$ & [-0.535, -0.071] & [0.434, 0.846] & [-0.563, -0.082] & [0.437, 0.840] & [-0.505, -0.059] & [0.470, 0.875] & [-0.665, -0.358] & [0.393, 0.606] & [-0.445, -0.147] & [0.595, 0.832] & [-0.578, -0.273] & [0.474, 0.697] \\
\cline{2-15}
 & \multirow[t]{2}{*}{Transparent Container} & meas. & -0.484 & 0.371 & -0.280 & 0.658 & -0.429 & 0.455 & -0.714 & 0.286 & -0.429 & 0.571 & -0.429 & 0.571 \\
 &  & $\text{CI}_{95\%}$ & [-0.791, -0.176] & [0.165, 0.837] & [-0.606, 0.046] & [0.392, 1.105] & [-0.734, -0.123] & [0.230, 0.900] & [-0.967, -0.462] & [0.132, 0.619] & [-0.691, -0.166] & [0.375, 0.871] & [-0.691, -0.166] & [0.375, 0.871] \\
\cline{2-15}
 & \multirow[t]{2}{*}{Preposition Replacement} & meas. & -0.769 & 0.000 & -0.518 & 0.367 & -0.586 & 0.255 & -0.700 & 0.300 & -0.600 & 0.400 & -0.750 & 0.250 \\
 &  & $\text{CI}_{95\%}$ & [-1.040, -0.498] & [0.000, nan] & [-0.854, -0.182] & [0.152, 0.884] & [-0.900, -0.271] & [0.076, 0.855] & [-0.982, -0.418] & [0.129, 0.699] & [-0.889, -0.311] & [0.207, 0.774] & [-1.046, -0.454] & [0.085, 0.732] \\
\cline{2-15}
 & \multirow[t]{2}{*}{Uninformative Label} & meas. & -0.769 & 0.000 & -0.568 & 0.306 & -0.536 & 0.318 & -0.875 & 0.125 & -0.625 & 0.375 & -0.375 & 0.625 \\
 &  & $\text{CI}_{95\%}$ & [-1.057, -0.482] & [0.000, nan] & [-0.916, -0.220] & [0.102, 0.919] & [-0.873, -0.198] & [0.106, 0.955] & [-1.151, -0.599] & [0.018, 0.891] & [-0.932, -0.318] & [0.176, 0.798] & [-0.682, -0.068] & [0.393, 0.994] \\
\cline{2-15}
 & \multirow[t]{2}{*}{Late Label} & meas. & -0.154 & 0.800 & -0.068 & 0.917 & -0.357 & 0.545 & -0.769 & 0.231 & -0.571 & 0.429 & -0.429 & 0.571 \\
 &  & $\text{CI}_{95\%}$ & [-0.473, 0.165] & [0.498, 1.284] & [-0.387, 0.251] & [0.610, 1.379] & [-0.666, -0.048] & [0.300, 0.991] & [-1.021, -0.518] & [0.089, 0.595] & [-0.834, -0.309] & [0.247, 0.743] & [-0.691, -0.166] & [0.375, 0.871] \\
\cline{2-15}
 & \multirow[t]{2}{*}{Non Protagonist Belief} & meas. & -0.055 & 0.929 & -0.040 & 0.951 & 0.060 & 1.077 & -0.500 & 0.500 & 0.000 & 1.000 & -0.143 & 0.857 \\
 &  & $\text{CI}_{95\%}$ & [-0.362, 0.252] & [0.613, 1.406] & [-0.377, 0.296] & [0.623, 1.451] & [-0.228, 0.349] & [0.755, 1.536] & [-0.764, -0.236] & [0.310, 0.808] & [-0.186, 0.186] & [0.830, 1.205] & [-0.374, 0.088] & [0.662, 1.110] \\
\cline{2-15}
 & \multirow[t]{2}{*}{Automatic Change Knowledge} & meas. & -0.369 & 0.520 & -0.568 & 0.306 & -0.486 & 0.382 & -0.333 & 0.667 & -0.500 & 0.500 & -0.500 & 0.500 \\
 &  & $\text{CI}_{95\%}$ & [-0.707, -0.031] & [0.257, 1.054] & [-0.916, -0.220] & [0.102, 0.919] & [-0.811, -0.160] & [0.159, 0.918] & [-0.663, -0.004] & [0.415, 1.070] & [-0.792, -0.208] & [0.292, 0.857] & [-0.792, -0.208] & [0.292, 0.857] \\
\cline{2-15}
 & \multirow[t]{2}{*}{Add Unrelated Information} & meas. & 0.088 & 1.114 & -0.461 & 0.437 & 0.214 & 1.273 & -0.357 & 0.643 & 0.000 & 1.000 & -0.143 & 0.857 \\
 &  & $\text{CI}_{95\%}$ & [-0.202, 0.377] & [0.778, 1.597] & [-0.779, -0.144] & [0.220, 0.868] & [-0.029, 0.458] & [0.950, 1.706] & [-0.616, -0.098] & [0.443, 0.934] & [-0.181, 0.181] & [0.834, 1.199] & [-0.374, 0.088] & [0.662, 1.110] \\
\cline{2-15}
 & \multirow[t]{2}{*}{Induction From Baseline} & meas. & -0.406 & 0.473 & -0.364 & 0.556 & -0.286 & 0.636 & -0.455 & 0.545 & -0.083 & 0.917 & -0.667 & 0.333 \\
 &  & $\text{CI}_{95\%}$ & [-0.735, -0.076] & [0.226, 0.988] & [-0.701, -0.026] & [0.299, 1.032] & [-0.606, 0.035] & [0.365, 1.110] & [-0.738, -0.171] & [0.337, 0.882] & [-0.312, 0.145] & [0.719, 1.169] & [-0.936, -0.397] & [0.162, 0.687] \\
\cline{2-15}
 & \multirow[t]{2}{*}{Untrustworthy Testimony} & meas. & -0.186 & 0.758 & -0.318 & 0.611 & -0.119 & 0.848 & -0.333 & 0.667 & -0.250 & 0.750 & -0.500 & 0.500 \\
 &  & $\text{CI}_{95\%}$ & [-0.511, 0.140] & [0.459, 1.254] & [-0.663, 0.026] & [0.338, 1.103] & [-0.434, 0.196] & [0.545, 1.321] & [-0.603, -0.064] & [0.457, 0.973] & [-0.511, 0.011] & [0.540, 1.042] & [-0.776, -0.224] & [0.301, 0.830] \\
\cline{2-15}
 & \multirow[t]{2}{*}{Conclusion From Sentiment} & meas. & -0.555 & 0.279 & -0.455 & 0.444 & -0.786 & 0.000 & -0.692 & 0.308 & -0.429 & 0.571 & -0.929 & 0.071 \\
 &  & $\text{CI}_{95\%}$ & [-0.854, -0.255] & [0.102, 0.762] & [-0.789, -0.121] & [0.213, 0.927] & [-1.029, -0.542] & [0.000, nan] & [-0.953, -0.431] & [0.145, 0.651] & [-0.691, -0.166] & [0.375, 0.871] & [-1.141, -0.716] & [0.007, 0.768] \\
\cline{1-15} \cline{2-15}
\end{longtable*}



\end{minipage}

}

\caption[Perturbation Effect Strengths: All Models. False-Belief-Tasks.]{Perturbation Effect Strengths: All Models. False-Belief-Tasks. ATE: Average Treatment; Effect RR: Relative Risk.}
\label{tab:RQ3:effect_strengths_pert_treatment_false_belief}
\end{table}
\end{landscape}

%% file: tables/RQ3/effect_strengths_cottreatment.tex
\begin{landscape}
\begin{table}[p]
\centering

{\tiny

\setlength{\tabcolsep}{2pt}  

\begin{minipage}{\linewidth}

\begin{longtable*}{llllllllllllll}
\toprule
 \textbf{ALL SUBTASKS} & Model & \multicolumn{2}{l}{meta-llama/Llama-2-70b-chat-hf} & \multicolumn{2}{l}{lmsys/vicuna-33b-v1.3} & \multicolumn{2}{l}{mistralai/Mixtral-8x7B-Instruct-v0.1} & \multicolumn{2}{l}{01-ai/Yi-34B-Chat} & \multicolumn{2}{l}{meta-llama/Meta-Llama-3-70B-Instruct} & \multicolumn{2}{l}{databricks/dbrx-instruct} \\
 & Effect & ATE & RR & ATE & RR & ATE & RR & ATE & RR & ATE & RR & ATE & RR \\
Perturbation Class & Statistic &  &  &  &  &  &  &  &  &  &  &  &  \\
\midrule
\endfirsthead
\toprule
 & Model & \multicolumn{2}{l}{meta-llama/Llama-2-70b-chat-hf} & \multicolumn{2}{l}{lmsys/vicuna-33b-v1.3} & \multicolumn{2}{l}{mistralai/Mixtral-8x7B-Instruct-v0.1} & \multicolumn{2}{l}{01-ai/Yi-34B-Chat} & \multicolumn{2}{l}{meta-llama/Meta-Llama-3-70B-Instruct} & \multicolumn{2}{l}{databricks/dbrx-instruct} \\
 & Effect & ATE & RR & ATE & RR & ATE & RR & ATE & RR & ATE & RR & ATE & RR \\
Perturbation Class & Statistic &  &  &  &  &  &  &  &  &  &  &  &  \\
\midrule
\endhead
\midrule
\multicolumn{14}{r}{Continued on next page} \\
\midrule
\endfoot
\bottomrule
\endlastfoot
\multirow[t]{2}{*}{no complication} & meas. & 0.000 & 1.000 & 0.045 & 1.053 & 0.098 & 1.122 & 0.067 & 1.082 & 0.000 & 1.000 & 0.010 & 1.011 \\
 & $\text{CI}_{95\%}$ & [-0.105, 0.105] & [0.880, 1.136] & [-0.058, 0.148] & [0.935, 1.187] & [0.000, 0.196] & [0.998, 1.261] & [-0.031, 0.165] & [0.964, 1.216] & [-0.061, 0.061] & [0.938, 1.066] & [-0.081, 0.101] & [0.912, 1.121] \\
\cline{1-14}
\multirow[t]{2}{*}{overall} & meas. & -0.020 & 0.975 & -0.016 & 0.981 & 0.068 & 1.095 & -0.021 & 0.973 & 0.024 & 1.029 & -0.003 & 0.996 \\
 & $\text{CI}_{95\%}$ & [-0.056, 0.017] & [0.930, 1.022] & [-0.053, 0.022] & [0.936, 1.027] & [0.031, 0.106] & [1.041, 1.152] & [-0.058, 0.016] & [0.927, 1.021] & [-0.005, 0.054] & [0.994, 1.065] & [-0.039, 0.033] & [0.951, 1.044] \\
\cline{1-14}
\multirow[t]{2}{*}{transparent container} & meas. & 0.048 & 1.060 & -0.211 & 0.753 & 0.129 & 1.181 & -0.030 & 0.963 & 0.010 & 1.011 & -0.019 & 0.978 \\
 & $\text{CI}_{95\%}$ & [-0.056, 0.152] & [0.933, 1.204] & [-0.333, -0.089] & [0.633, 0.896] & [0.016, 0.242] & [1.017, 1.371] & [-0.140, 0.080] & [0.840, 1.105] & [-0.079, 0.098] & [0.915, 1.116] & [-0.117, 0.079] & [0.872, 1.096] \\
\cline{1-14}
\multirow[t]{2}{*}{preposition replacement} & meas. & -0.012 & 0.982 & -0.118 & 0.855 & 0.167 & 1.351 & 0.000 & 1.000 & 0.051 & 1.066 & 0.043 & 1.061 \\
 & $\text{CI}_{95\%}$ & [-0.153, 0.128] & [0.802, 1.203] & [-0.260, 0.025] & [0.704, 1.038] & [0.016, 0.317] & [1.021, 1.788] & [-0.136, 0.136] & [0.834, 1.199] & [-0.075, 0.176] & [0.910, 1.247] & [-0.103, 0.189] & [0.866, 1.300] \\
\cline{1-14}
\multirow[t]{2}{*}{uninformative label} & meas. & -0.109 & 0.844 & 0.067 & 1.089 & -0.047 & 0.942 & 0.031 & 1.045 & 0.033 & 1.042 & 0.078 & 1.104 \\
 & $\text{CI}_{95\%}$ & [-0.270, 0.051] & [0.657, 1.086] & [-0.080, 0.213] & [0.902, 1.314] & [-0.188, 0.094] & [0.788, 1.127] & [-0.124, 0.186] & [0.838, 1.304] & [-0.106, 0.173] & [0.878, 1.236] & [-0.062, 0.219] & [0.923, 1.322] \\
\cline{1-14}
\multirow[t]{2}{*}{late label} & meas. & 0.040 & 1.052 & 0.011 & 1.014 & -0.020 & 0.974 & -0.119 & 0.850 & 0.010 & 1.011 & -0.029 & 0.965 \\
 & $\text{CI}_{95\%}$ & [-0.074, 0.153] & [0.910, 1.217] & [-0.106, 0.127] & [0.873, 1.176] & [-0.137, 0.097] & [0.836, 1.136] & [-0.239, 0.001] & [0.719, 1.005] & [-0.086, 0.106] & [0.905, 1.131] & [-0.137, 0.079] & [0.843, 1.104] \\
\cline{1-14}
\multirow[t]{2}{*}{non protagonist belief} & meas. & 0.048 & 1.058 & 0.000 & 1.000 & 0.082 & 1.107 & -0.058 & 0.930 & 0.029 & 1.033 & 0.038 & 1.046 \\
 & $\text{CI}_{95\%}$ & [-0.050, 0.146] & [0.942, 1.188] & [-0.092, 0.092] & [0.903, 1.108] & [-0.029, 0.192] & [0.964, 1.271] & [-0.167, 0.051] & [0.811, 1.067] & [-0.054, 0.113] & [0.942, 1.133] & [-0.059, 0.135] & [0.934, 1.172] \\
\cline{1-14}
\multirow[t]{2}{*}{automatic change knowledge} & meas. & -0.075 & 0.893 & 0.016 & 1.024 & -0.037 & 0.936 & -0.118 & 0.806 & -0.038 & 0.949 & -0.113 & 0.820 \\
 & $\text{CI}_{95\%}$ & [-0.218, 0.068] & [0.718, 1.110] & [-0.144, 0.175] & [0.805, 1.303] & [-0.187, 0.112] & [0.719, 1.219] & [-0.303, 0.067] & [0.572, 1.137] & [-0.175, 0.098] & [0.788, 1.143] & [-0.262, 0.037] & [0.628, 1.071] \\
\cline{1-14}
\multirow[t]{2}{*}{add unrelated information} & meas. & -0.048 & 0.944 & -0.055 & 0.935 & 0.178 & 1.231 & -0.029 & 0.965 & -0.010 & 0.990 & 0.019 & 1.023 \\
 & $\text{CI}_{95\%}$ & [-0.150, 0.054] & [0.834, 1.068] & [-0.167, 0.057] & [0.815, 1.073] & [0.084, 0.273] & [1.095, 1.383] & [-0.135, 0.078] & [0.846, 1.100] & [-0.073, 0.054] & [0.926, 1.058] & [-0.080, 0.119] & [0.910, 1.151] \\
\cline{1-14}
\multirow[t]{2}{*}{induction from baseline} & meas. & -0.092 & 0.892 & 0.049 & 1.059 & 0.114 & 1.164 & 0.024 & 1.030 & 0.080 & 1.095 & -0.011 & 0.984 \\
 & $\text{CI}_{95\%}$ & [-0.209, 0.025] & [0.769, 1.034] & [-0.061, 0.159] & [0.930, 1.205] & [-0.012, 0.240] & [0.981, 1.381] & [-0.095, 0.143] & [0.889, 1.193] & [-0.019, 0.178] & [0.977, 1.226] & [-0.144, 0.121] & [0.817, 1.186] \\
\cline{1-14}
\multirow[t]{2}{*}{untrustworthy testimony} & meas. & 0.031 & 1.043 & 0.047 & 1.063 & 0.083 & 1.116 & 0.042 & 1.056 & 0.128 & 1.164 & 0.052 & 1.068 \\
 & $\text{CI}_{95\%}$ & [-0.091, 0.154] & [0.885, 1.229] & [-0.081, 0.174] & [0.898, 1.259] & [-0.036, 0.203] & [0.952, 1.308] & [-0.078, 0.162] & [0.902, 1.237] & [0.023, 0.232] & [1.024, 1.324] & [-0.063, 0.168] & [0.922, 1.238] \\
\cline{1-14}
\multirow[t]{2}{*}{conclusion from sentiment} & meas. & -0.098 & 0.878 & 0.000 & 1.000 & -0.029 & 0.959 & -0.058 & 0.927 & -0.010 & 0.988 & -0.077 & 0.886 \\
 & $\text{CI}_{95\%}$ & [-0.215, 0.019] & [0.751, 1.027] & [-0.125, 0.125] & [0.855, 1.170] & [-0.153, 0.095] & [0.802, 1.147] & [-0.173, 0.058] & [0.796, 1.080] & [-0.118, 0.099] & [0.863, 1.131] & [-0.205, 0.051] & [0.722, 1.086] \\
\cline{1-14}
\end{longtable*}


\begin{longtable*}{llllllllllllll}
\toprule
 \textbf{FALSE BELIEF SUBTASKS} & Model & \multicolumn{2}{l}{meta-llama/Llama-2-70b-chat-hf} & \multicolumn{2}{l}{lmsys/vicuna-33b-v1.3} & \multicolumn{2}{l}{mistralai/Mixtral-8x7B-Instruct-v0.1} & \multicolumn{2}{l}{01-ai/Yi-34B-Chat} & \multicolumn{2}{l}{meta-llama/Meta-Llama-3-70B-Instruct} & \multicolumn{2}{l}{databricks/dbrx-instruct} \\
 & Effect & ATE & RR & ATE & RR & ATE & RR & ATE & RR & ATE & RR & ATE & RR \\
Perturbation Class & Statistic &  &  &  &  &  &  &  &  &  &  &  &  \\
\midrule
\endfirsthead
\toprule
 & Model & \multicolumn{2}{l}{meta-llama/Llama-2-70b-chat-hf} & \multicolumn{2}{l}{lmsys/vicuna-33b-v1.3} & \multicolumn{2}{l}{mistralai/Mixtral-8x7B-Instruct-v0.1} & \multicolumn{2}{l}{01-ai/Yi-34B-Chat} & \multicolumn{2}{l}{meta-llama/Meta-Llama-3-70B-Instruct} & \multicolumn{2}{l}{databricks/dbrx-instruct} \\
 & Effect & ATE & RR & ATE & RR & ATE & RR & ATE & RR & ATE & RR & ATE & RR \\
Perturbation Class & Statistic &  &  &  &  &  &  &  &  &  &  &  &  \\
\midrule
\endhead
\midrule
\multicolumn{14}{r}{Continued on next page} \\
\midrule
\endfoot
\bottomrule
\endlastfoot
\multirow[t]{2}{*}{no complication} & meas. & 0.154 & 1.250 & 0.545 & 3.000 & -0.143 & 0.846 & 0.571 & 2.333 & 0.000 & 1.000 & 0.214 & 1.273 \\
 & $\text{CI}_{95\%}$ & [-0.165, 0.473] & [0.779, 2.006] & [0.218, 0.873] & [1.200, 7.502] & [-0.410, 0.125] & [0.614, 1.166] & [0.309, 0.834] & [1.345, 4.047] & [-0.181, 0.181] & [0.834, 1.199] & [-0.029, 0.458] & [0.950, 1.706] \\
\cline{1-14}
\multirow[t]{2}{*}{overall} & meas. & 0.000 & 1.000 & 0.068 & 1.160 & 0.030 & 1.062 & 0.023 & 1.050 & 0.081 & 1.131 & 0.037 & 1.069 \\
 & $\text{CI}_{95\%}$ & [-0.118, 0.118] & [0.776, 1.288] & [-0.057, 0.194] & [0.882, 1.525] & [-0.088, 0.147] & [0.835, 1.352] & [-0.097, 0.143] & [0.816, 1.351] & [-0.030, 0.193] & [0.955, 1.339] & [-0.080, 0.155] & [0.866, 1.321] \\
\cline{1-14}
\multirow[t]{2}{*}{transparent container} & meas. & 0.000 & 1.000 & -0.154 & 0.778 & 0.286 & 5.000 & -0.143 & 0.667 & 0.143 & 1.333 & 0.143 & 1.333 \\
 & $\text{CI}_{95\%}$ & [-0.308, 0.308] & [0.340, 2.939] & [-0.482, 0.175] & [0.449, 1.347] & [0.004, 0.568] & [0.430, 58.188] & [-0.459, 0.174] & [0.263, 1.692] & [-0.182, 0.468] & [0.683, 2.604] & [-0.182, 0.468] & [0.683, 2.604] \\
\cline{1-14}
\multirow[t]{2}{*}{preposition replacement} & meas. & -0.200 & 1.000 & -0.100 & 0.750 & 0.000 & 1.000 & -0.300 & 0.500 & 0.200 & 2.000 & -0.125 & 0.667 \\
 & $\text{CI}_{95\%}$ & [-0.487, 0.087] & [0.000, inf] & [-0.461, 0.261] & [0.260, 2.161] & [-0.335, 0.335] & [0.188, 5.330] & [-0.661, 0.061] & [0.195, 1.282] & [-0.151, 0.551] & [0.519, 7.708] & [-0.511, 0.261] & [0.182, 2.448] \\
\cline{1-14}
\multirow[t]{2}{*}{uninformative label} & meas. & -0.125 & 1.000 & -0.125 & 0.667 & -0.250 & 0.500 & -0.125 & 0.500 & 0.125 & 1.500 & 0.000 & 1.000 \\
 & $\text{CI}_{95\%}$ & [-0.434, 0.184] & [0.000, inf] & [-0.511, 0.261] & [0.182, 2.448] & [-0.639, 0.139] & [0.149, 1.673] & [-0.487, 0.237] & [0.054, 4.657] & [-0.261, 0.511] & [0.409, 5.507] & [-0.394, 0.394] & [0.532, 1.880] \\
\cline{1-14}
\multirow[t]{2}{*}{late label} & meas. & 0.077 & 1.143 & 0.417 & 2.250 & 0.000 & 1.000 & -0.154 & 0.600 & 0.000 & 1.000 & -0.071 & 0.889 \\
 & $\text{CI}_{95\%}$ & [-0.256, 0.410] & [0.639, 2.045] & [0.088, 0.745] & [1.038, 4.876] & [-0.325, 0.325] & [0.469, 2.133] & [-0.473, 0.165] & [0.196, 1.836] & [-0.325, 0.325] & [0.469, 2.133] & [-0.393, 0.250] & [0.522, 1.515] \\
\cline{1-14}
\multirow[t]{2}{*}{non protagonist belief} & meas. & 0.071 & 1.111 & 0.111 & 1.167 & 0.154 & 1.222 & -0.071 & 0.875 & 0.077 & 1.083 & 0.071 & 1.091 \\
 & $\text{CI}_{95\%}$ & [-0.242, 0.385] & [0.698, 1.768] & [-0.253, 0.476] & [0.700, 1.945] & [-0.150, 0.458] & [0.815, 1.833] & [-0.397, 0.254] & [0.474, 1.614] & [-0.147, 0.301] & [0.855, 1.372] & [-0.211, 0.354] & [0.772, 1.542] \\
\cline{1-14}
\multirow[t]{2}{*}{automatic change knowledge} & meas. & -0.200 & 0.667 & -0.125 & 0.667 & -0.300 & 0.500 & 0.000 & 1.000 & 0.000 & 1.000 & -0.100 & 0.833 \\
 & $\text{CI}_{95\%}$ & [-0.567, 0.167] & [0.306, 1.453] & [-0.511, 0.261] & [0.182, 2.448] & [-0.661, 0.061] & [0.195, 1.282] & [-0.429, 0.429] & [0.525, 1.904] & [-0.370, 0.370] & [0.477, 2.098] & [-0.469, 0.269] & [0.423, 1.643] \\
\cline{1-14}
\multirow[t]{2}{*}{add unrelated information} & meas. & 0.286 & 1.500 & -0.143 & 0.714 & 0.357 & 1.556 & 0.000 & 1.000 & 0.000 & 1.000 & 0.214 & 1.333 \\
 & $\text{CI}_{95\%}$ & [-0.014, 0.585] & [0.947, 2.376] & [-0.465, 0.180] & [0.327, 1.561] & [0.098, 0.616] & [1.071, 2.259] & [-0.318, 0.318] & [0.609, 1.641] & [-0.181, 0.181] & [0.834, 1.199] & [-0.082, 0.510] & [0.880, 2.021] \\
\cline{1-14}
\multirow[t]{2}{*}{induction from baseline} & meas. & -0.182 & 0.667 & -0.182 & 0.714 & 0.000 & 1.000 & 0.182 & 1.500 & 0.250 & 1.375 & 0.000 & 1.000 \\
 & $\text{CI}_{95\%}$ & [-0.536, 0.172] & [0.292, 1.520] & [-0.536, 0.172] & [0.362, 1.408] & [-0.346, 0.346] & [0.500, 2.000] & [-0.172, 0.536] & [0.658, 3.420] & [-0.054, 0.554] & [0.911, 2.075] & [-0.335, 0.335] & [0.366, 2.736] \\
\cline{1-14}
\multirow[t]{2}{*}{untrustworthy testimony} & meas. & 0.250 & 1.750 & 0.500 & 1.000 & 0.167 & 1.333 & 0.333 & 2.000 & 0.083 & 1.125 & 0.167 & 1.500 \\
 & $\text{CI}_{95\%}$ & [-0.090, 0.590] & [0.767, 3.992] & [0.191, 0.809] & [1.000, 1.000] & [-0.174, 0.508] & [0.728, 2.443] & [-0.002, 0.669] & [0.903, 4.432] & [-0.245, 0.412] & [0.705, 1.795] & [-0.174, 0.508] & [0.632, 3.559] \\
\cline{1-14}
\multirow[t]{2}{*}{conclusion from sentiment} & meas. & -0.286 & 0.429 & 0.000 & 1.000 & -0.143 & 1.000 & -0.154 & 0.667 & 0.071 & 1.143 & -0.214 & 0.250 \\
 & $\text{CI}_{95\%}$ & [-0.596, 0.024] & [0.147, 1.250] & [-0.351, 0.351] & [0.381, 2.623] & [-0.374, 0.088] & [0.000, inf] & [-0.482, 0.175] & [0.271, 1.639] & [-0.254, 0.397] & [0.620, 2.108] & [-0.490, 0.062] & [0.021, 3.019] \\
\cline{1-14}
\end{longtable*}


\end{minipage}

}

\caption[Chain of Thought Effect Strengths: All Models and Perturbation Classes.]{Chain of Thought Effect Strengths: All Models and Perturbation Classes. Upper table created using "All-Subtasks". Lower table measurements on "False Belief" Tasks. ATE: Average Treatment Effect; RR: Relative Risk.}

\label{tab:RQ3:effect_strengths_cottreatment}
\end{table}
\end{landscape}